\documentclass[journal]{IEEEtran}
\usepackage{graphicx,float,stfloats,amsthm,amssymb,multirow,pifont,amsmath,scalerel,makecell}
\usepackage{bm,hyperref,xcolor,enumitem,booktabs,arydshln,footnote,url}
\usepackage[caption=false]{subfig}
\usepackage{color}
\usepackage{mathrsfs}
\usepackage{algorithm,algorithmic}
\usepackage{threeparttable}
\usepackage[subfigure,titles]{tocloft}
\usepackage[comma, numbers]{natbib}
\hypersetup{hypertex=true,
	colorlinks=true,
	linkcolor=blue,
	anchorcolor=blue,
	citecolor=blue}

\begin{document}
\title{A Unified Framework with Multimodal Fine-tuning for Remote Sensing Semantic Segmentation}
\author{Xianping~Ma,
				Xiaokang~Zhang,~\IEEEmembership{Senior Member,~IEEE,}
				Man-On~Pun,~\IEEEmembership{Senior Member,~IEEE,}
				and~Bo Huang
		\thanks{This work was supported in part by the National Natural Science Foundation of China under Grant 42371374 and 41801323, China Postdoctoral Science Foundation under grant 2020M682038 and the Shenzhen Science and Technology Innovation Committee under Grant No. JCYJ20190813170803617. \textit{(Corresponding authors: Man-On Pun, Xiaokang Zhang)}}
		\thanks{Xianping Ma, and Man-On Pun are with the School of Science and Engineering, The Chinese University of Hong Kong, Shenzhen, Shenzhen 518172, China (e-mails: xianpingma@link.cuhk.edu.cn; SimonPun@cuhk.edu.cn).}
		\thanks{Xiaokang Zhang is with the School of Information Science and Engineering, Wuhan University of Science and Technology, Wuhan 430081, China.(e-mail: natezhangxk@gmail.com).}
		\thanks{Bo Huang is with the Department of Geography, The University of Hong Kong, Hong Kong, SAR 999077, China (e-mail: hbcuhk@gmail.com).}}
\maketitle

\begin{abstract}
Multimodal remote sensing data, acquired from diverse sensors, offer a comprehensive and integrated perspective of the Earth's surface. Leveraging multimodal fusion techniques, semantic segmentation enables detailed and accurate analysis of geographic scenes, surpassing single-modality approaches. Building on advancements in vision foundation models, particularly the Segment Anything Model (SAM), this study proposes a unified framework incorporating a novel Multimodal Fine-tuning Network (MFNet) for remote sensing semantic segmentation. The proposed framework is designed to seamlessly integrate with various fine-tuning mechanisms, demonstrated through the inclusion of Adapter and Low-Rank Adaptation (LoRA) as representative examples. This extensibility ensures the framework's adaptability to other emerging fine-tuning strategies, allowing models to retain SAM's general knowledge while effectively leveraging multimodal data. Additionally, a pyramid-based Deep Fusion Module (DFM) is introduced to integrate high-level geographic features across multiple scales, enhancing feature representation prior to decoding. This work also highlights SAM's robust generalization capabilities with Digital Surface Model (DSM) data, a novel application. Extensive experiments on three benchmark multimodal remote sensing datasets, ISPRS Vaihingen, ISPRS Potsdam and MMHunan, demonstrate that the proposed MFNet significantly outperforms existing methods in multimodal semantic segmentation, setting a new standard in the field while offering a versatile foundation for future research and applications. The source code for this work is accessible at \href{https://github.com/sstary/SSRS}{https://github.com/sstary/SSRS}.
\end{abstract}

\begin{IEEEkeywords}
Multimodal Fine-tuning, Remote Sensing, Semantic Segmentation, SAM, Adapter, LoRA
\end{IEEEkeywords}

\IEEEpeerreviewmaketitle

\section{Introduction}\label{sec:int}
Multimodal remote sensing semantic segmentation involves the process of classifying each pixel in ground images using data from multiple sources or modalities, such as optical images, multispectral images, hyperspectral images, and LiDAR. By integrating diverse types of information, multimodal approaches enhance the accuracy and robustness of segmentation, particularly in complex environments \citep{gomez2015multimodal, li2022deep}. This technique leverages the complementary strengths of different data types to improve the identification of land objects, making it crucial for applications such as land use and land cover \citep{yao2023extended, hong2021multimodal}, environmental monitoring \citep{karmakar2023crop}, and disaster management \citep{algiriyage2022multi, zhang2023cross}. In recent years, deep learning technologies have greatly prompted the development of multimodal fusion methods in remote sensing.

Initially, convolution neural networks (CNNs) were the dominant architecture, known for their ability to extract local spatial features from different modalities with the encoder-decoder framework \citep{UNet, li2021abcnet}. These early CNN-based methods stack multimodal data and fuse features at various stages for improved segmentation performance \citep{FuseNet, zhang2022multilevel}. With the introduction of the self-attention-based Transformer \citep{tf}, which excels at modeling global context and long-range dependencies, many hybrid methods integrating Transformers into CNN-based methods have emerged. In particular, Vision Transformer (ViT) \citep{vit} and Swin Transformer \citep{liu2021swin} further introduced Transformer to the computer vision community, greatly improving the ability of image feature extraction. Combining CNNs for detailed feature extraction with Transformers for capturing global relationships between different data sources marked a new stage in segmentation models \citep{ma2022crossmodal, he2023mftransnet, ma2024frequency}. This hybrid approach enhances the ability to fuse information across modalities and scales, leading to more accurate, robust segmentation results in complex scenes. Despite their good performance, the aforementioned models were trained solely on narrowly scoped task-specific data, limiting their acquisition of general visual knowledge. In contrast, foundation models benefit from large-scale, diverse pretraining data, enabling them to develop broad visual representations that extend beyond single-task constraints.

\begin{figure}[h]
	\centering
	{\includegraphics[width=0.7\linewidth]{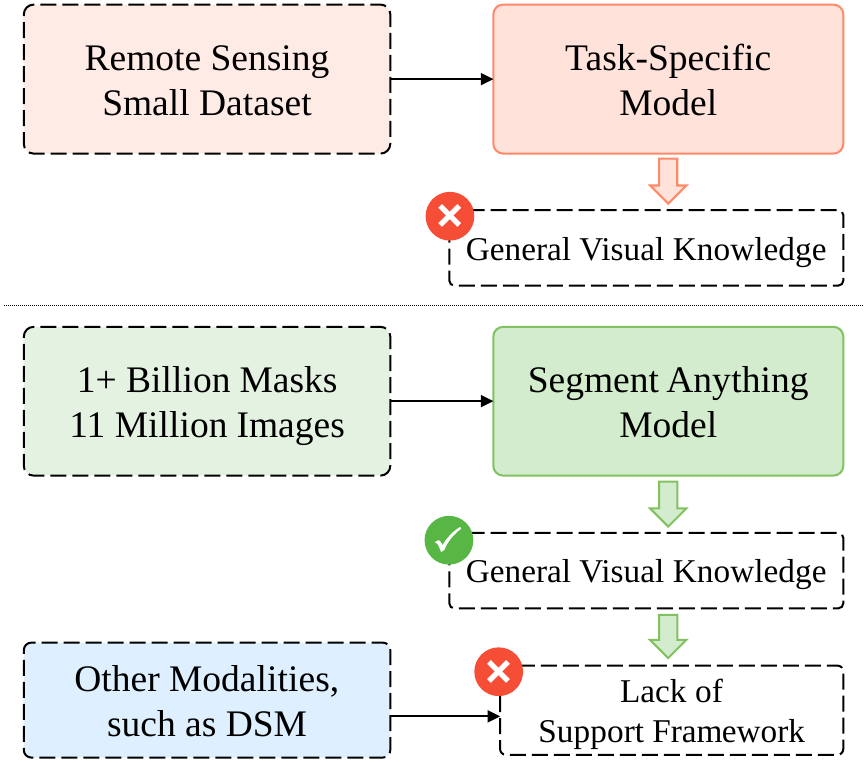}}
	\caption{Illustration of the key challenges: Traditional task-specific remote sensing models are limited in general visual knowledge. In contrast, SAM contains general knowledge from large-scale natural image corpora, but it lacks the framework to support multimodal remote sensing tasks.}
	\label{fig_challenge}
\end{figure}

Interestingly, the recent emergence of large foundation models offers a solution to this challenge. In particular, SAM \citep{kirillov2023segment} is a cutting-edge segmentation model designed to tackle a wide range of segmentation tasks across diverse datasets. It is comprised of three components, including a ViT-based image encoder, a prompt encoder and a mask decoder. Developed by Meta AI, SAM benefits from large-scale training on a vast visual corpus of natural images, enabling it to generalize effectively to unseen objects. Its versatility and robust performance position it as a valuable tool for various applications \citep{ma2024segment, wang2024sam}. However, compared to natural images, remote sensing images exhibit significant differences due to variations in sensors, resolution, spectral ranges and so on \citep{li2018deep, zheng2020foreground, ma2023unsupervised}. For multimodal tasks, non-optical data such as DSM further increases the discrepancy. This raises a new challenge: \emph{How can the capabilities of SAM, acquired from massive natural visual corpora, be leveraged to enhance multimodal remote sensing tasks?} Fig.~\ref{fig_challenge} illustrates the key challenges discussed above, providing a clearer depiction of the problem addressed in this work.

Existing fine-tuning techniques, especially Adapter \citep{houlsby2019parameter, chen2022adaptformer, he2023parameter} and Low-Rank Adaptation (LoRA) \citep{hu2021lora, zhang2023adaptive}, address part of this challenge. Compared to training and full fine-tuning, these methods fix most of the parameters and learn task-specific information with very few parameters. This approach enables parameter-efficient learning and the successful migration of large foundation models to a wider range of specific downstream tasks, even in more constrained hardware environments. The concept of Adapter was first introduced in the natural language processing community \citep{houlsby2019parameter} as a method to fine-tune large pretrained models for specific downstream tasks. The core idea behind Adapter is to introduce a parallel, compact, and scalable adaptation module that learns task-specific knowledge during training while the original model branch remains fixed, retaining task-agnostic knowledge. Similarly, LoRA \citep{hu2021lora} introduced trainable rank decomposition matrices to learn task-specific knowledge. This approach leverages the synergy between task-specific and task-agnostic knowledge, enabling efficient fine-tuning of the large model.

\begin{figure}[t]
	\centering
	{\includegraphics[width=\linewidth]{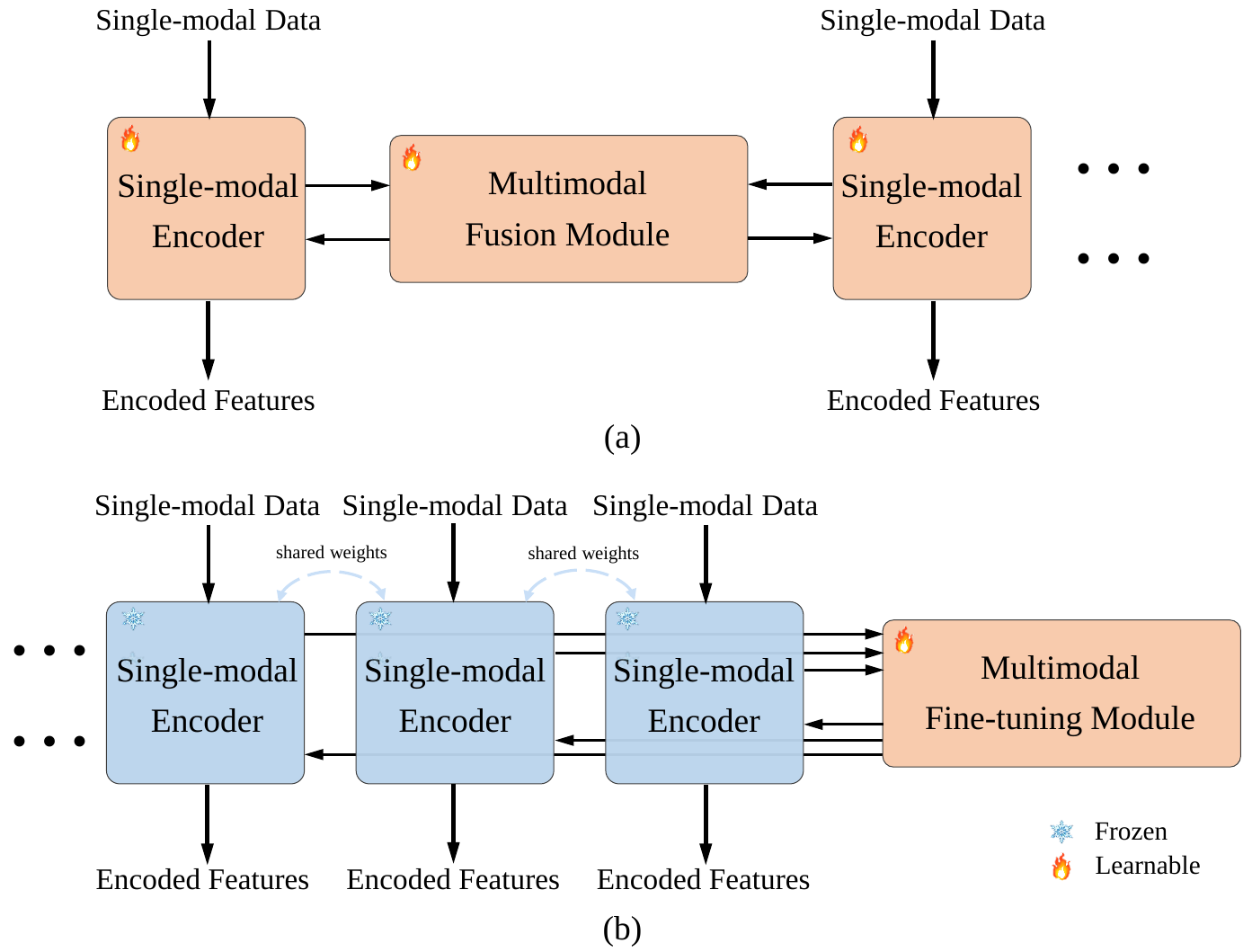}}
	\caption{(a) Current multimodal fusion methods, and (b) The proposed unified multimodal fine-tuning framework. In our method, single-modal encoders are used to learn general knowledge and frozen, enabling multimodal tasks to benefit from the vision foundational model.}
	\label{fig_unify}
\end{figure}

In the field of remote sensing, fine-tuning techniques have focused on adapting SAM for single-modality tasks \citep{pu2024classwise, zhou2024mesam, lu2024multi}. For instance, CWSAM \citep{pu2024classwise} and MeSAM \citep{zhou2024mesam} adapted SAM's image encoder and introduced custom mask decoders for remote sensing data. However, an analysis of the SAM parameter distribution across three scales reveals that the majority of SAM's parameters are concentrated in its image encoder, suggesting that most of its general knowledge is encapsulated within this component. Therefore, we consider it unnecessary to modify SAM's prompt encoder and mask decoder for remote sensing-specific tasks, as such modifications increase model adaptation complexity while hindering seamless integration with existing segmentation models. 

To address the aforementioned challenges with fine-tuning framework, we present a unified multimodal fine-tuning framework for multimodal data feature learning and fusion. Specifically, as illustrated in Fig.~\ref{fig_unify}(a), current multimodal fusion methods typically assign separate encoders for each modality and perform feature fusion through a multimodal fusion module \citep{ResUNet-a, ma2024ftransunet, zhang2024asanet}. When additional modalities are introduced, more encoders need to be added accordingly. During training, all encoders and the fusion module need to be trained simultaneously. In contrast, the unified multimodal fine-tuning framework proposed in this work revolutionizes this structure. As depicted in Fig.~\ref{fig_unify}(b), it capitalizes on the general knowledge embedded in the foundational model and facilitates modality learning and fusion through the multimodal fine-tuning module. During training, the encoders remain fixed, requiring optimization only for the multimodal fine-tuning module. This approach not only fully leverages the vision foundational model but also offers seamless scalability to incorporate additional modalities.

After that, to thoroughly investigate the impact of various fine-tuning mechanisms in remote sensing, we introduce two multimodal fine-tuning techniques, namely Multimodal Adapter (MMAdapter) and Multimodal LoRA (MMLoRA), for multimodal fusion remote sensing tasks. It is worth noting that our multimodal fine-tuning framework represents a novel feature learning fusion strategy that operates independently of the specific fine-tuning module used. Leveraging MMAdapter and MMLoRA, respectively, we propose a novel multimodal fusion method, namely Multimodal Fine-tuning Network (MFNet) with different image encoder architectures. Furthermore, MFNet utilizes a Deep Fusion Module (DFM) to perform multiscale processing and fusion of the deep high-level features from SAM's image encoder. This is an effective solution for the complex characteristics of remote sensing data. Additionally, we employ a universal semantic segmentation decoder \citep{wang2022unetformer} that does not require additional task-specific design efforts, allowing for easy integration with decoders from other tasks. The fourfold contributions are summarized in the following:
\begin{itemize}[leftmargin=*]
\item A unified multimodal fine-tuning framework is proposed to adapt SAM for multimodal feature learning, independent of the specific design of the fine-tuning module and the number of data modalities;

\item A scalable multimodal fusion network, namely MFNet, is proposed by capitalizing on SAM's image encoder and the proposed multimodal fine-tuning framework to perform remote sensing semantic segmentation. This is a streamlined and flexible adaptation network that eliminates most of the redundant modules in SAM;

\item Two most representative fine-tuning architectures from the literature, namely Adapter and LoRA, are utilized to validate the framework's effectiveness by showing SAM's robust generalization capability on DSM for the first time. Extensive experiments on three well-known multimodal remote sensing datasets, ISPRS Vaihingen, ISPRS Potsdam and MMHunan, confirm that the proposed MFNet substantially outperforms existing methods in terms of semantic segmentation performance;

\item To the best of our knowledge, as pioneers in exploring the multimodal fine-tuning approach based on SAM, this work thoroughly investigates the role of the two most widely used fine-tuning mechanisms in the field of remote sensing. It establishes a robust foundation for related studies and provides clear directions for future research.
\end{itemize}

The remainder of this paper is structured as follows. Sec.~\ref{sec:rel} provides a review of related works on multimodal fusion and SAM. Sec.~\ref{sec:met} presents the unified multimodal fine-tuning framework and MFNet, followed by a detailed description and analysis of the extensive experiments and a discussion in Sec.~\ref{sec:exp}. Finally, the conclusion is given in Sec.~\ref{sec:con}.

\section{Related works}\label{sec:rel}
\subsection{Multimodal Remote Sensing Semantic Segmentation}
Semantic segmentation is a critical preprocessing step in remote sensing image processing, and leveraging multimodal information often yields better results than relying on a single modality. Recently, the advent of deep learning has revolutionized the entire field of remote sensing, including semantic segmentation. Based on the classical encoder-decoder framework \citep{UNet, li2021abcnet}, numerous multimodal fusion approaches based on CNNs and Transformers have driven significant advancements in the field \citep{vfusenet, ma2022crossmodal, he2023mftransnet, ma2024ftransunet}. ResUNet-a \citep{ResUNet-a}, an early CNN-based architecture, simply stacked multimodal data into four channels. Furthermore, vFuseNet \citep{vfusenet} introduced a two-branch encoder to separately extract multimodal features, enabling deeper multimodal fusion through element-wise operations at the feature level. Recently, the introduction of Transformers \citep{tf, vit} has further enriched multimodal networks. For instance, CMFNet \citep{ma2022crossmodal} employed CNNs for feature extraction and uses a Transformer structure to connect multimodal features across scales, emphasizing the importance of scale in multimodal fusion. Similarly, MFTransNet \citep{ma2024ftransunet} used CNNs for feature extraction, while enhancing the self-attention module with spatial and channel attention for better feature fusion. FTransUNet \citep{ma2024ftransunet} presented a multilevel fusion approach to refine the fusion of shallow- and deep-level remote sensing semantic features. Despite the great performance they achieved, we argue that existing models lack sufficient general knowledge, which poses a fundamental limitation to the advancement of multimodal fusion methods. 

\begin{figure*}[t]
	\centering
	{\includegraphics[width=0.95\linewidth]{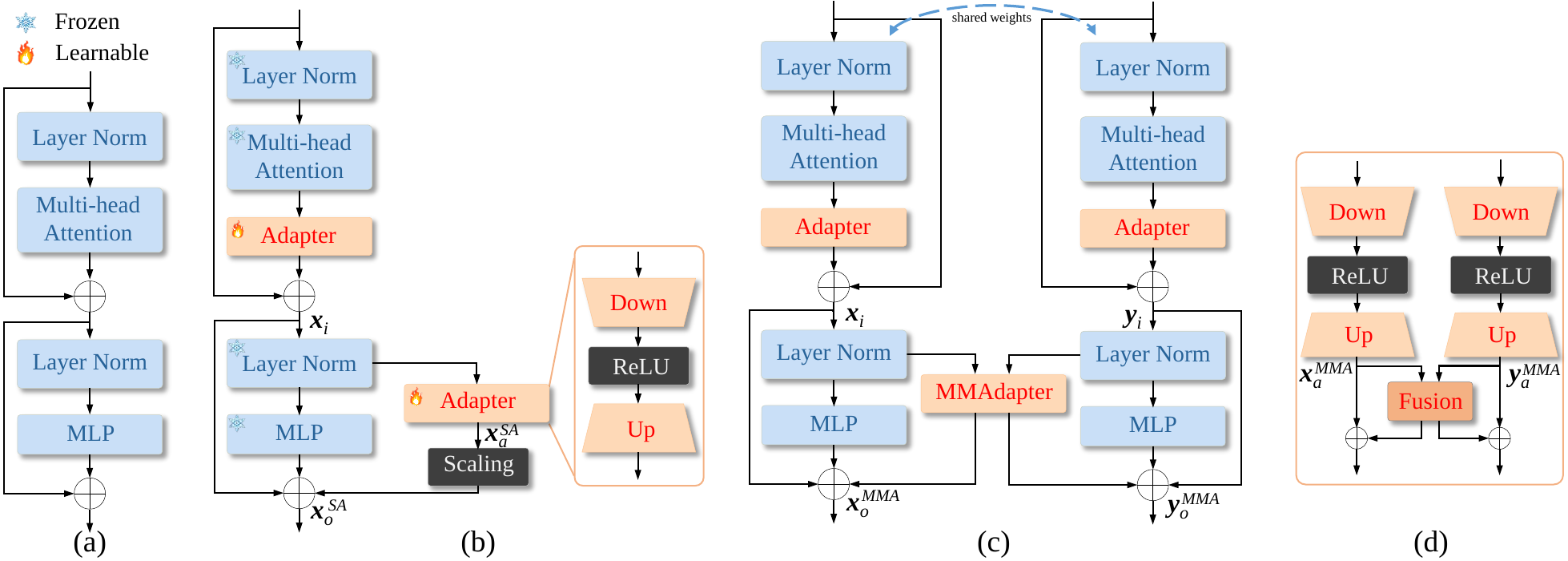}}
	\caption{(a) The ViT block {\em without} Adapter in SAM's image encoder, (b) the ViT block equipped with standard Adapter \citep{wu2023medical}, (c) the ViT block endowed with the proposed MMAdapter and (d) the detailed structure of the MMAdapter. The Adapter facilitates the efficient utilization of general knowledge in specific tasks. Compared to standard Adapter, MMAadapter is characterized by two branches of shared weights for multimodal feature extraction. The standard Adapter and the proposed MMAdapter are used to fine-tune and fuse features, respectively.}
	\label{fig_mmadapter}
\end{figure*}

\subsection{SAM in Remote Sensing}
The SAM \citep{kirillov2023segment} holds a unique position as a general-purpose image segmentation model. This vision foundation model was trained on a very large visual corpus. It endows SAM with a remarkable ability to generalize to unseen objects, making it well-suited for applications in diverse scenarios. Nowadays, SAM has already been applied across various fields, such as autonomous driving \citep{li2024fusionsam}, medical image processing \citep{mazurowski2023segment} and remote sensing \citep{SAMRS, qi2024samp, chen2024change, ma2024sam}. In remote sensing, SAMRS \citep{SAMRS} leveraged SAM to integrate numerous existing remote sensing datasets, using a new prompt named rotated bounding box. Furthermore, some recent works have considered fine-tuning SAM for remote sensing tasks such as semantic segmentation \citep{pu2024classwise, zhou2024mesam, yan2023ringmo} and change detection \citep{ding2023adapting, mei2024scd}.

For single modality tasks, CWSAM \citep{pu2024classwise} adapted SAM to synthetic aperture radar (SAR) images by introducing a task-specific input module and a class-wise mask decoder. MeSAM \citep{zhou2024mesam} incorporated an inception mixer into the SAM's image encoder to retain high-frequency features and introduced a multiscale-connected mask decoder for optical images. SAM\_MLoRA \citep{lu2024multi} employed multiple LoRA modules in parallel to enhance LoRA's decomposition capability. For multimodal tasks, RingMo-SAM \citep{yan2023ringmo} introduced a prompt encoder tailored to multimodal remote sensing data, along with a category-decoupling mask decoder. It is observed that these methods focus on refining the fine-tuning mechanisms and designing task-specific prompts or mask decoders. They have preliminarily explored the generalization ability of SAM in remote sensing tasks. However, as discussed in Sec.~\ref{sec:int}, the general knowledge in SAM is primarily centered in the image encoder. While these approaches successfully utilize SAM's general knowledge to remote sensing, their complex architectures greatly hinder their adaptation to existing general semantic segmentation networks. Furthermore, there is currently {\em no} SAM-based multimodal approach designed for DSM data.

\section{Proposed Method}\label{sec:met}
We first introduce the unified multimodal fine-tuning framework by elaborating the MMAdapter and MMLoRA. Specifically, we review the conventional single-modal fine-tuning strategy Adapter and the proposed MMAdapter in Sec.~\ref{sec:mmadapter}. After that, we present another classic single-modal fine-tuning strategy LoRA and the proposed MMLoRA in Sec.~\ref{sec:mmlora}. Building upon the proposed multimodal fine-tuning mechanisms, the proposed MFNet is elaborated in Sec.~\ref{sec:mfnet}. Notably, MFNet has two distinct architectures based on the selection between MMAdapter and MMLoRA. Finally, to provide a clear explanation, we use the example of two modalities for illustration.

\subsection{Standard Adapter and the Proposed MMAdapter}\label{sec:mmadapter}
\begin{itemize}[leftmargin=*]
\item{\bf Standard Adapter}: As discussed in Sec.~\ref{sec:int}, the general knowledge is confined to SAM's image encoders, specifically the ViT blocks, whose structure is shown in Fig.\ref{fig_mmadapter}(a). In \citep{wu2023medical}, the Adapter was proposed to enhance the capabilities of the ViT blocks for medical tasks through fine-tuning, as illustrated in Fig.~\ref{fig_mmadapter}(b). Instead of adjusting all parameters, the pretrained SAM parameters remain frozen, while two Adapter modules are introduced to learn task-specific knowledge. Each Adapter consists of a down projection, a ReLU activation, and an up projection. The down projection compresses the input embedding to a lower dimension using a simple MLP layer, and the up projection restores the compressed embedding to its original dimension using another MLP layer. For a specific input feature $\bm{x}_{i}\in \mathbb{R}^{h\times w \times c}$, where $h$, $w$, and $c$ represent the height, width, and channels of the input feature, the Adapter's process for generating the adapted feature can be expressed as:
\begin{align}
\bm{x}^{SA}_{a}&=\mathrm{ReLU}\left(\mathrm{LN}(\bm{x}_{i})\cdot \bm{W}_{d}\right) \cdot \bm{W}_{u},
\end{align}
where $\bm{W}_{d}\in \mathbb{R}^{c\times \hat{c}}$ and $\bm{W}_{u}\in \mathbb{R}^{\hat{c}\times c}$ are the down projection and up projection, respectively. $\hat{c}\ll c$ is the compressed middle dimension of the Adapter. After that, both the adapted feature $\bm{x}_{a}$ and the output of the original MLP branch are fused with $\bm{x}_{i}$ by residual connection to generate the output feature $\bm{x}_{o}$:
\begin{align}
\bm{x}^{SA}_{o}&=\mathcal{F}(\bm{x}_{i})+s\cdot\bm{x}^{SA}_{a}+\bm{x}_{i},
\end{align}
where $\mathcal{F}(\cdot)$ denotes the MLP operation and $s$ is a scaling factor to weight the task-specific and task-agnostic knowledge. Since the Adapter proposed in \citep{wu2023medical} was designed for single-modal data, it is referred to as the standard Adapter in the sequel.

\begin{figure*}[t]
	\centering
	{\includegraphics[width=0.7\linewidth]{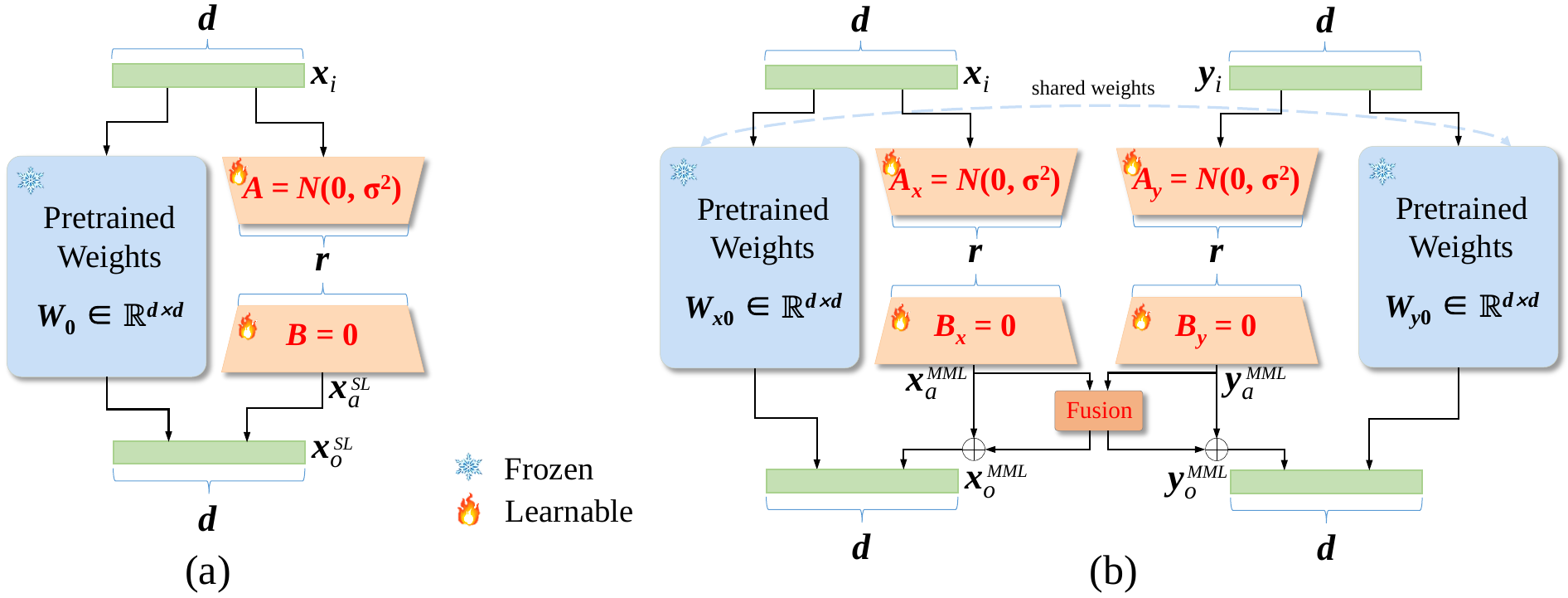}}
	\caption{(a) The detailed structure of standard LoRA \citep{hu2021lora} and (b) the proposed MMLoRA. It is observed that MMLoRA adopts the design principles of MMAdapter, which not only reduces the module's complexity but also highlights the versatility of this strategy.}
	\label{fig_mmlora}
\end{figure*}

\item{\bf Proposed MMAdapter}: Next, we extend the standard Adapter to multimodal tasks. The proposed MMAdapter is a core component in the proposed multimodal fine-tuning framework. As illustrated in Fig.~\ref{fig_mmadapter}(c), we employ dual branches with shared weights to process multimodal information. The Adapter after the Multi-head Attention is retained to extract features from each modality independently, while the Adapter during the MLP stage is replaced with the proposed MMAdapter. The detail of MMAdapter is presented in Fig.~\ref{fig_mmadapter}(d). While preserving the core structure of the Adapter, the MMAdapter enables modality interaction through a fusion module. Notably, this design accommodates {\em arbitrary} feature fusion strategies. To emphasize the effectiveness of the multimodal fine-tuning framework, we adopt the simplest method of element-wise addition based on two weighting factors, $\lambda_{1}$ and $\lambda_{2}$. For two specific multimodal input features, $\bm{x}_{i}\in \mathbb{R}^{h\times w \times c}$ and $\bm{y}_{i}\in \mathbb{R}^{h\times w \times c}$, the process of generating the adapted features with MMAdapter can be described as:
\begin{align}
\bm{x}^{MMA}_{a}&=\mathrm{ReLU}\left(\mathrm{LN}(\bm{x}_{i})\cdot \bm{W}_{dx}\right) \cdot \bm{W}_{ux},\\
\bm{y}^{MMA}_{a}&=\mathrm{ReLU}\left(\mathrm{LN}(\bm{y}_{i})\cdot \bm{W}_{dy}\right) \cdot \bm{W}_{uy},
\end{align}
where $\bm{W}_{dx}, \bm{W}_{dy}\in \mathbb{R}^{c\times \hat{c}}$ and $\bm{W}_{ux}, \bm{W}_{uy}\in \mathbb{R}^{\hat{c}\times c}$ are the down projections and up projections, respectively. After that, the multimodal output features $\bm{x}^{MMA}_{o}$ and $\bm{y}^{MMA}_{o}$ are generated using $\lambda_{1}$ and $\lambda_{2}$ as follows:
\begin{align}
\bm{x}^{MMA}_{o}&=\mathcal{F}(\bm{x}_{i})+\lambda_{1}\cdot\bm{x}^{MMA}_{a}+(1-\lambda_{1})\cdot\bm{y}^{MMA}_{a}+\bm{x}_{i},\\
\bm{y}^{MMA}_{o}&=\mathcal{F}(\bm{y}_{i})+\lambda_{2}\cdot\bm{y}^{MMA}_{a}+(1-\lambda_{2})\cdot\bm{x}^{MMA}_{a}+\bm{y}_{i}.
\end{align}

Only newly added parameters are optimized, while other parameters remain fixed during fine-tuning. Detailed annotations are provided in Fig.~\ref{fig_mmadapter}(b), while other subfigures omit these annotations for clarity and conciseness.
\end{itemize}

\subsection{Standard LoRA and the Proposed MMLoRA}\label{sec:mmlora}
\begin{itemize}[leftmargin=*]
\item{\bf Standard LoRA}:
Foundation models are composed of numerous dense layers, typically utilizing full-rank matrix multiplication. To adapt these pretrained models for specific tasks, LoRA \citep{hu2021lora} assumes that during the adaptation process, the updates to the weights have a lower ``intrinsic rank" \citep{aghajanyan2020intrinsic}. This mechanism can be applied to any linear layer. For a pretrained weight matrix $\bm{W}_{0} \in \mathbb{R}^{d \times d}$, the update is expressed with a low-rank decomposition: 
\begin{equation}
	\bm{W}_{0} + \Delta\bm{W} = \bm{W}_{0} + \bm{B}\bm{A},
\end{equation}
where $\bm{B} \in \mathbb{R}^{d \times r}$, $\bm{A} \in \mathbb{R}^{r \times d}$, and the rank $r \ll d$. 

During training, $\bm{W}_{0}$ remains fixed and does not receive gradient updates, while the trainable parameters are contained in $\bm{A}$ and $\bm{B}$. Given an input $\bm{x}_{i}$, the forward computation for the adapted module is represented as:
\begin{equation}
\bm{x}^{SL}_{o}=(\bm{W}_{0}+\Delta\bm{W})\bm{x}_{i}=\bm{W}_{0}\bm{x}_{i}+\bm{x}^{SL}_{a}.
\end{equation}

The matrix $\bm{A}$ is initialized with a random Gaussian distribution, while $\bm{B}$ is initialized to $0$, resulting in $\Delta\bm{W} = 0$ at the start of training. The architecture of LoRA is depicted in Fig.~\ref{fig_mmlora}(a). Throughout this work, the single-modal implementation of LoRA is referred to as the standard LoRA.

\item{\bf Proposed MMLoRA}: Similar to MMAdapter, we extend standard LoRA to handle multimodal tasks. As shown in Fig.~\ref{fig_mmlora}(b), a dual-branch structure with shared weights is employed to process multimodal information. This design enables the learning of task-specific knowledge both within individual modalities and across modalities, facilitated by the fusion module. Given inputs $\bm{x}_{i}$ and $\bm{y}_{i}$, the process of generating the adapted features using MMLoRA can be described as:
\begin{align}
\bm{x}^{MML}_{o}&=\bm{W}_{x0}\bm{x}_{i}+\lambda_{1}\cdot\bm{x}^{MML}_{a}+(1-\lambda_{1})\bm{y}^{MML}_{a},\\
\bm{y}^{MML}_{o}&=\bm{W}_{y0}\bm{y}_{i}+\lambda_{2}\cdot\bm{y}^{MML}_{a}+(1-\lambda_{2})\bm{x}^{MML}_{a},
\end{align}
where
\begin{align}
\bm{x}^{MML}_{a}&=\bm{B}_{x}\bm{A}_{x}\bm{x}_{i},\\
\bm{y}^{MML}_{a}&=\bm{B}_{y}\bm{A}_{y}\bm{y}_{i}.
\end{align}

Finally, it is worth mentioning that the design shown in Fig.~\ref{fig_mmadapter} and Fig.~\ref{fig_mmlora} can be generalized to more than two modalities in a straightforward manner.
\end{itemize}

\begin{figure*}[t]
	\centering
	{\includegraphics[width=\linewidth]{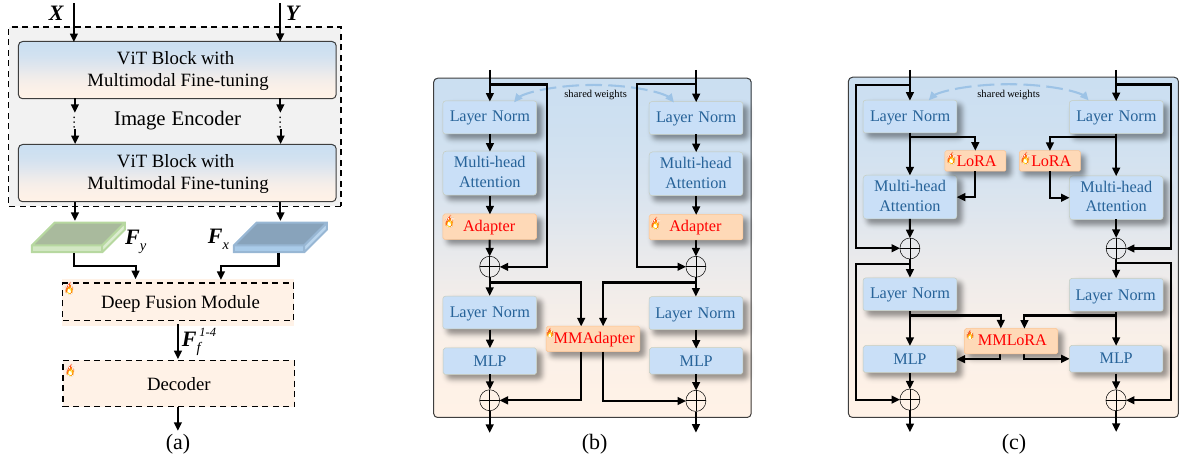}}
	\caption{(a) Overview of the proposed MFNet consisting of a SAM's image encoder with multimodal fine-tuning, a DFM and a general Decoder. The structure of (b) ViT block with MMAdapter and (c) ViT block with MMLoRA. These constitute the two distinct architectures of MFNet.}
	\label{fig_framework}
\end{figure*}

\subsection{The Proposed MFNet}\label{sec:mfnet}
Fig.~\ref{fig_framework} shows an overview of the proposed MFNet and two different multimodal fune-tuning strategies. The input to MFNet is first processed by SAM's image encoder endowed with MMAdapter or MMLoRA, which is responsible for extracting and fusing multimodal remote sensing features using the multimodal fine-tuning mechanism. The output is then fed into DFM, which receives two single-scale multimodal outputs from the encoder and expands them into two sets of multiscale multimodal features using pyramid modules. These high-level abstract features are then fused by four squeeze-and-excitation (SE) Fusion modules to generate a group of multiscale features. Finally, the outputs of DFM are passed to the decoder to produce segmentation prediction maps. In this section, we introduce the key components of the proposed MFNet in detail.

\begin{itemize}[leftmargin=*]
\item {\bf SAM's image encoder}: We denote the optical images and their corresponding DSM data as $\bm{X}\in \mathbb{R}^{H\times W \times 3}$ and $\bm{Y}\in \mathbb{R}^{H\times W \times 1}$, respectively, where $H$ and $W$ represent the height and width of the inputs. The SAM's image encoder, employing a non-hierarchical ViT backbone, first embeds the input into a size of $\mathbb{R}^{h\times w \times c}$, where $h=\frac{H}{16}$, $w=\frac{W}{16}$, and $c$ is the embedding dimension. Next, stacked ViT Blocks are used to extract features whose size is maintained throughout the encoding process \citep{li2022exploring}. As illustrated in Fig.~\ref{fig_framework}(a), both $\bm{X}$ and $\bm{Y}$ are input into the SAM's image encoder. It is important to note that the same SAM encoder is used for DSM data, which demonstrates that SAM can be utilized to extract features from non-image data. The SAM's image encoder extracts and fuses multimodal features, generating high-level abstract features $\bm{F}_{x}\in \mathbb{R}^{h\times w \times c}$ and $\bm{F}_{y}\in \mathbb{R}^{h\times w \times c}$ through the multimodal fine-tuning modules.

\item {\bf ViT block with MMLoRA}: Fig.\ref{fig_framework}(b) depicts the architecture of MMAdapter within the ViT block. Conversely, MMLoRA serves as a multimodal fine-tuning method applied in parallel with linear layers. For greater clarity, the structure of the ViT block incorporating MMLoRA is presented in Fig.\ref{fig_framework}(c). In the multi-head attention module, LoRA module is applied to the $q$ and $v$ projection layers \citep{zhang2023customized}. At this stage, multimodal interaction is excluded to concentrate on capturing task-specific information within each modality. In the subsequent MLP layer, MMLoRA module is applied to the linear layers in MLP, facilitating the fusion of multimodal information.

\item{\bf DFM}: Multiscale features play a critical role in semantic segmentation tasks since dense predicting requires both global and local information. As shown in Fig.~\ref{fig_modules}(a), two pyramid modules, each consisting of a set of parallel convolutions or deconvolutions, are used to generate multiscale feature maps. Starting with the plain ViT feature map at a scale of $\frac{1}{16}$, we produce feature maps at scales of $\{\frac{1}{4}, \frac{1}{8}, \frac{1}{16}, \frac{1}{32}\}$ using convolutions with strides of $\{\frac{1}{4}, \frac{1}{2}, 1, 2\}$, where fractional strides indicate deconvolutions \citep{li2022exploring}. These simple pyramid modules generate two sets of multimodal multiscale features, denoted as $\bm{F}_{x}^{i}$ and $\bm{F}_{y}^{i}$, where $i=\{1,2,3,4\}$ represents the scale index. Subsequently, four SE Fusion modules \citep{ma2024ftransunet} are employed to further integrate the multimodal features. It is worth mentioning that more advanced fusion modules can yield further improved segmentation performance.

\begin{figure}[t]
	\centering
	{\includegraphics[width=1\linewidth]{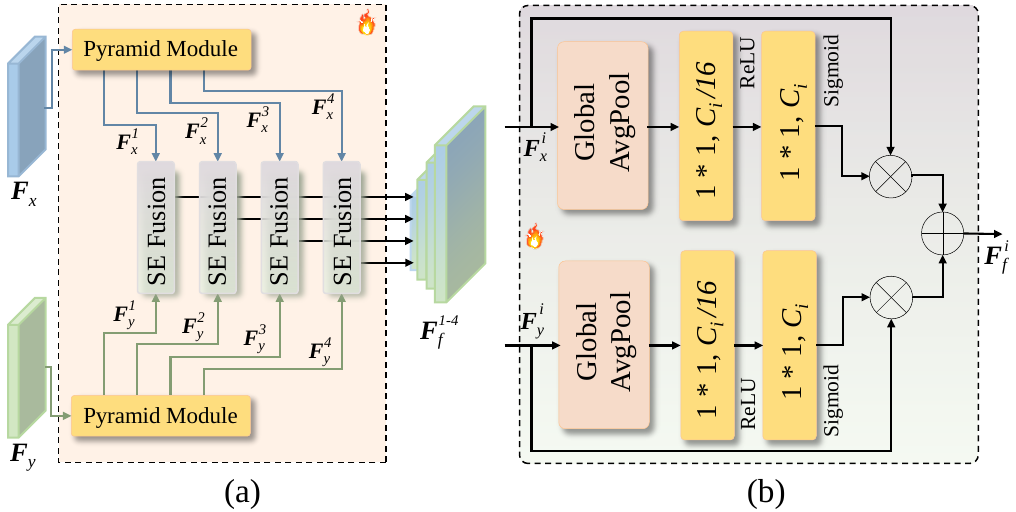}}
	\caption{(a) Schematic of the DFM. There are two Pyramid modules that expand multimodal features for multiscale features before they are fused by four SE Fusion modules. (b) Schematic of the SE Fusion module \citep{ma2024ftransunet}. Notably, we just employ the existing simple fusion module and do not design this structure specifically, which proves that our primary gains are derived from the multimodal fine-tuning strategy based on the vision foundation model.}
	\label{fig_modules}
\end{figure}

As illustrated in Fig.~\ref{fig_modules}(b), the SE Fusion module begins by aggregating the global information from the multimodal features. For the $i$-th fusion module, with an input channel size of $C_{i}$, the squeeze and excitation process is performed through two convolutional operations with a kernel size of $1 \times 1$, followed by the ReLU and Sigmoid activations. The multimodal outputs are then weighted and combined element-wise, producing the enhanced fused features denoted by $\bm{F}_{f}^{i}$. The outputs from the four SE Fusion modules form the multiscale fusion features, denoted as $\bm{F}_{f}^\emph{1-4}$, which are fed into the decoder for processing. The decoder introduced in UNetformer \citep{wang2022unetformer} is employed in this work that reconstructs abstract semantic information into a segmented map by focusing on both global and local information.
\end{itemize}

\section{Experiments and Discussion}\label{sec:exp}
\subsection{Datasets}
\textbf{Vaihingen:} It contains $16$ fine-resolution Orthophotos, each averaging $2500\times2000$ pixels. These Orthophotos consist of three channels: Near-Infrared, Red, and Green (NIRRG), and come with a normalized DSM at a $9$ cm ground sampling distance. The $16$ Orthophotos are split into a training set of $12$ patches and a test set of $4$ patches. To improve the storage and reading efficiency of large patches, a sliding window of size $256 \times 256$ is utilized rather than cropping the patches into smaller images in both training and test stages, which results in about $960$ training images and $320$ test images.

\textbf{Potsdam:} This dataset is much larger than the Vaihingen dataset, which consists of $24$ high-resolution Orthophotos, each with the size of $6000\times 6000$ pixels. It includes four multispectral bands: Infrared, Red, Green, and Blue (IRRGB), along with a normalized DSM of $5$ cm. The last three bands (RGB) in this dataset are utilized to diversify our experiments. The $24$ orthophotos are split into $18$ patches for training and $6$ for testing. Using the same sliding window approach, this dataset contains $10368$ training samples and $3456$ test samples.

The Vaihingen and Potsdam datasets classify five main foreground categories: Building (Bui.), Tree (Tre.), Low Vegetation (Low.), Car, and Impervious Surface (Imp.). Additionally, a background class labeled as Clutter contains indistinguishable debris and water surfaces. Notably, the sliding window to collect training samples moves with a smaller step size, and the overlapping areas are averaged to the reduced boundary effect during the test stage.

\begin{figure}[bp!]
\centering
{\includegraphics[width=\linewidth]{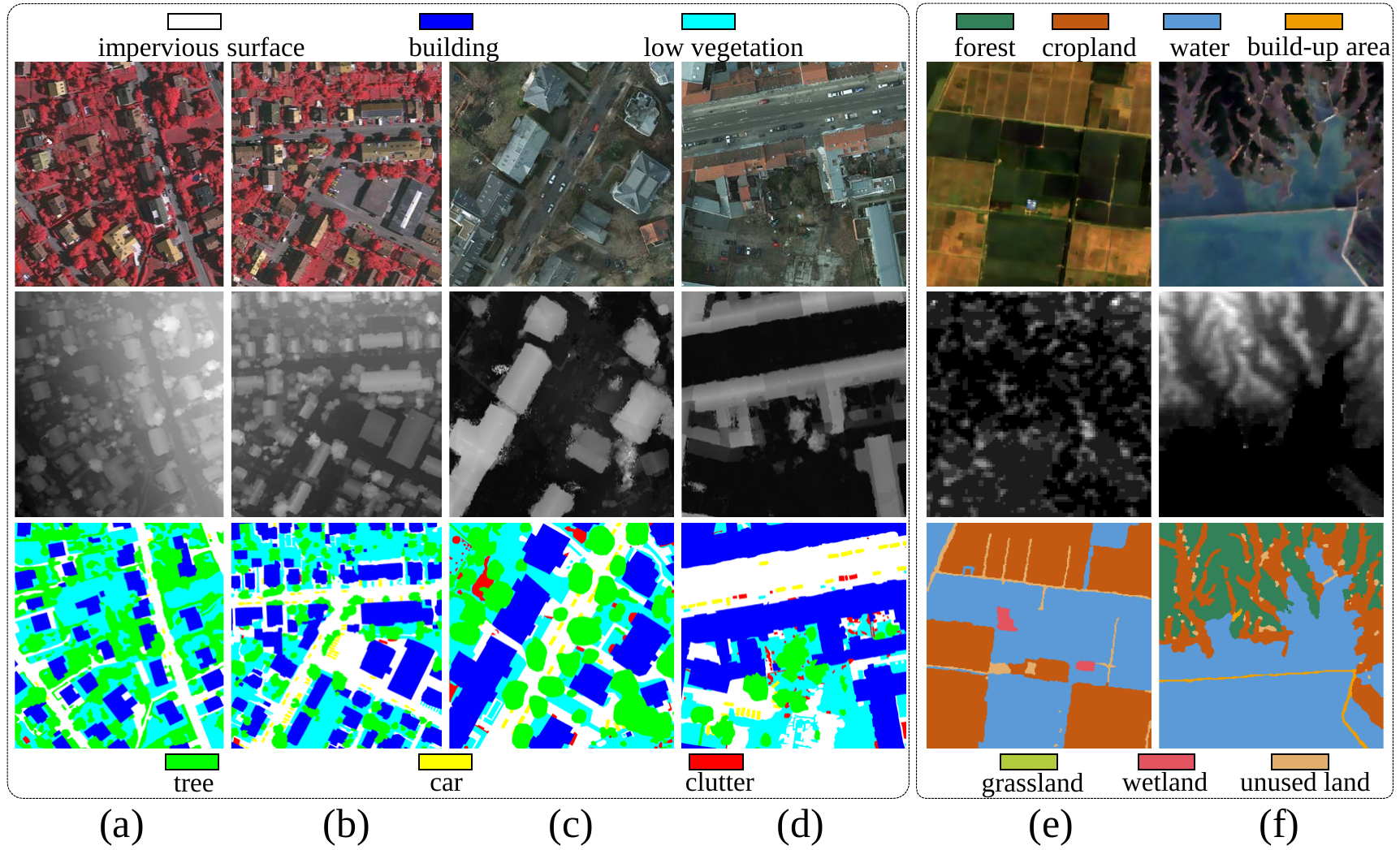}}
\caption{Here we present (a/b): two samples of size $2048 \times 2048$ from Vaihingen, (c/d): two samples of size $2048 \times 2048$ from Potsdam (last two columns), and  (e/f): two samples of size $256 \times 256$ from MMHunan. The first row displays the orthophotos with NIRRG channels for Vaihingen, RGB channels for Potsdam, and RGB channels for MMHunan. The second and third rows present the corresponding pixel-wise depth information and ground truth labels. They show the individual and complementary characteristics of remote sensing data from different sources.}
\label{data_show}
\end{figure}

\textbf{MMHunan:} This dataset \citep{li2022dkdfn} differs significantly in spatial resolution from that of the ISPRS datasets, with a value of $10$ meters. It contains $500$ Sentinel-2 image patches, each of size $256 \times 256$ pixels, accompanied by corresponding Digital Elevation Data. We selected the Red, Green, and Blue bands to form the visible imagery. The dataset includes annotations for $7$ land cover types: Cropland, Forest, Grassland, Wetland, Water, Unused Land, and Built-up Area.

Visual examples from all three datasets are presented in Fig.~\ref{data_show}. The substantial differences in both data characteristics and land cover categories greatly enhance the diversity of our experiments.

\begin{table*}[t]\scriptsize
	\centering
	\caption{Quantitative results on the Vaihingen dataset. The best results are in \textbf{Bold}. The second best results are \underline{underlined}. (\%).}
		\begin{tabular}{cccccccccc}
			\hline
      \multirow{2}{*}{\textbf{Method}} & \multirow{2}{*}{\textbf{Backbone}} & \multicolumn{6}{c}{\textbf{OA}} & \multirow{2}{*}{\textbf{mF1}} & \multirow{2}{*}{\textbf{mIoU}} \\ \cline{3-8}
      & & \textbf{Bui.}~ & \textbf{Tre.}  & \textbf{Low.}  & \textbf{Car}   & \textbf{Imp.}  & \textbf{Total}  &                      &                       \\ \cline{1-1} \cline{9-10} 
			\hline
			FuseNet \citep{FuseNet}  & VGG16 & 96.28 & 90.28 & 78.98 & 81.37 & 91.66 & 90.51 & 87.71 & 78.71  \\
			vFuseNet \citep{vfusenet} & VGG16 & 95.92 & 91.36 & 77.64 & 76.06 & 91.85 & 90.49 & 87.89 & 78.92  \\
			MAResU-Net \citep{MAResUNet} & ResNet18 & 94.84 & 89.99 & 79.09 & 85.89 & 92.19 & 90.17 & 88.54 & 79.89  \\
			ESANet \citep{ESANet}   & ResNet34 & 95.69 & 90.50 & 77.16 & 85.46 & 91.39 & 90.61 & 88.18 & 79.42  \\
			CMGFNet \citep{CMGFNet} & ResNet34 & 97.75 & 91.60 & 80.03 & 87.28 & 92.35 & 91.72 & 90.00 & 82.26  \\
			PSPNet \citep{PSPNet}  & ResNet101 & 94.52 & 90.17 & 78.84 & 79.22 & 92.03 & 89.94 & 86.55 & 76.96  \\
		  SA-GATE \citep{SAGATE}  & ResNet101 & 94.84 & 92.56 & 81.29 & 87.79 & 91.69 & 91.10 & 89.81 & 81.27  \\
			\hline
			CMFNet \citep{ma2022crossmodal}  & VGG16 & 97.17 & 90.82 & 80.37 & 85.47 & 92.36 & 91.40 & 89.48 & 81.44  \\
			UNetFormer \citep{wang2022unetformer} & ResNet18 & 96.23 & 91.85 & 79.95 & 86.99 & 91.85 & 91.17 & 89.48 & 81.97  \\
			MFTransNet \citep{he2023mftransnet} & ResNet34 & 96.41 & 91.48 & 80.09 & 86.52 & 92.11 & 91.22 & 89.62 & 81.61  \\
			TransUNet \citep{Transunet} & R50-ViT-B & 96.48 & 92.77 & 76.14 & 69.56 & 91.66 & 90.96 & 87.34 & 78.26  \\
			FTransUNet \citep{ma2024ftransunet} & R50-ViT-B & 98.20 & 91.94 & 81.49 & \textbf{91.27} & 93.01 & 92.40 & 91.21    & 84.23 \\
			RS$^{3}$Mamba \citep{ma2024rs} & R18-Mamba-T & 97.40 & 92.14 & 79.56 & 88.15 & 92.19 & 91.64 & 90.34 & 82.78 \\
			FTransDeepLab \citep{FTransDeepLab} & ResNet101 & 98.11 & 93.45 & 80.35 & 89.98 & 93.23 & 92.61 & 91.00 & 83.87 \\
			MultiSenseSeg \citep{MultiSenseSeg} & Segformer-B2 & 97.91 & 93.04 & 81.58 & 89.06 & \underline{93.56} & 92.73 & 91.42 & 84.53 \\
			\hline
			\multirow{3}{*}{MFNet (MMLoRA)}  & ViT-B & 97.83 & \textbf{94.26} & 77.82 & 85.43 & 91.98 & 91.93 & 89.89 & 82.09 \\
			  & ViT-L & 96.85 & 92.89 & 81.09 & 89.95 & 93.28 & 92.22 & 91.09 & 83.96 \\
				& ViT-H & 97.98 & 92.35 & \underline{82.96} & 90.09 & 93.25 & 92.73 & 91.50 & 84.66 \\
			\hline
			\multirow{3}{*}{MFNet (MMAdapter)}  & ViT-B & \underline{98.73} & 91.41 & \textbf{83.09} & 85.63 & 92.91 & 92.62 & 90.60 & 83.24 \\
				& ViT-L & \textbf{98.84} & 93.17 & 81.16 & 89.23 & 93.39 & \underline{92.93} & \underline{91.51} & \underline{84.72} \\
			  & ViT-H & 98.38 & \underline{93.94} & 80.70 & \underline{90.47} & \textbf{93.59} & \textbf{92.97} & \textbf{91.71} & \textbf{85.03} \\
			\hline	
	\end{tabular}\label{tab:vlist}
\end{table*}

\begin{table*}[t]\scriptsize
	\centering
	\caption{Quantitative results on the Potsdam dataset. The best results are in \textbf{Bold}. The second best results are \underline{underlined}. (\%).}
		\begin{tabular}{cccccccccc}
			\hline
      \multirow{2}{*}{\textbf{Method}} & \multirow{2}{*}{\textbf{Backbone}} & \multicolumn{6}{c}{\textbf{OA}} & \multirow{2}{*}{\textbf{mF1}} & \multirow{2}{*}{\textbf{mIoU}} \\ \cline{3-8}
      & & \textbf{Bui.}~ & \textbf{Tre.}  & \textbf{Low.}  & \textbf{Car}   & \textbf{Imp.}  & \textbf{Total}  &                      &                       \\ \cline{1-1} \cline{9-10} 
			\hline
			FuseNet \citep{FuseNet} & VGG16 & 97.48 & 85.14 & 87.31 & 96.10 & 92.64 & 90.58 & 91.60 & 84.86  \\
			vFuseNet \citep{vfusenet}  & VGG16 & 97.23 & 84.29 & 89.03 & 95.49 & 91.62 & 90.22 & 91.26 & 84.26  \\
			MAResU-Net \citep{MAResUNet} & ResNet18 & 96.82 & 83.97 & 87.70 & 95.88 & 92.19 & 89.82 & 90.86 & 83.61  \\
			ESANet \citep{ESANet}   & ResNet34 & 97.10 & 85.31 & 87.81 & 94.08 & 92.76 & 89.74 & 91.22 & 84.15  \\
			CMGFNet \citep{CMGFNet} & ResNet34 & 97.41 & 86.80 & 86.68 & 95.68 & 92.60 & 90.21 & 91.40 & 84.53  \\
			PSPNet \citep{PSPNet} & ResNet101 & 97.03 & 83.13 & 85.67 & 88.81 & 90.91 & 88.67 & 88.92 & 80.36  \\
			SA-GATE \citep{SAGATE}  & ResNet101 & 96.54 & 81.18 & 85.35 & \underline{96.63} & 90.77 & 87.91 & 90.26 & 82.53  \\
			\hline
			CMFNet \citep{ma2022crossmodal}  & VGG16 & 97.63 & 87.40 & 88.00 & 95.68 & 92.84 & 91.16 & 92.10 & 85.63  \\
			UNetFormer \citep{wang2022unetformer}   & ResNet18 & 97.69  & 86.47 & 87.93 & 95.91 & 92.27 & 90.65 & 91.71 & 85.05  \\
			MFTransNet \citep{he2023mftransnet}  & ResNet34 & 97.37 & 85.71 & 86.92 & 96.05 & 92.45 & 89.96 & 91.11 & 84.04  \\
			TransUNet \citep{Transunet}  & R50-ViT-B & 96.63 & 82.65 & 89.98 & 93.17 & 91.93 & 90.01 & 90.97 & 83.74  \\
			FTransUNet \citep{ma2024ftransunet} & R50-ViT-B & 97.78 & 88.27 & 88.48 & 96.31 & \underline{93.17} & 91.34 & 92.41 & 86.20 \\
			RS$^{3}$Mamba \citep{ma2024rs} & R18-Mamba-T & 97.70 & 86.11 & 89.53 & 96.23 & 91.36 & 90.49 & 91.69 & 85.01 \\
			FTransDeepLab \citep{FTransDeepLab} & ResNet101 & 97.58 & 85.87 & \underline{90.08} & \textbf{96.94} & 92.81 & 90.97 & 92.08 & 85.62 \\
			MultiSenseSeg \citep{MultiSenseSeg} & Segformer-B2 & \underline{98.32} & 87.65 & 89.54 & 96.27 & 92.46 & 91.30 & 92.35 & 86.10 \\
			\hline
			\multirow{3}{*}{MFNet (MMLoRA)}  & ViT-B & 97.60 & 86.45 & 87.87 & 94.39 & 92.44 & 90.57 & 91.48 & 84.61 \\
			  & ViT-L & 97.59 & \underline{88.57} & 88.34 & 96.35 & 92.68 & 90.99 & 92.13 & 85.71 \\
				& ViT-H & 98.19 & 87.30 & 89.89 & 96.27 & 92.80 & 91.43 & 92.49 & 86.34 \\
			\hline
			\multirow{3}{*}{MFNet (MMAdapter)}  & ViT-B & 97.93 & 87.13 & 87.72 & 95.68 & 92.68 & 90.89 & 91.79 & 85.14 \\
			  & ViT-L & 98.31 & \textbf{88.78} & 87.27 & 96.29 & \textbf{93.69} & \underline{91.62} & \underline{92.51} & \underline{86.37} \\
				& ViT-H & \textbf{98.44} & 87.37 & \textbf{90.36} & 96.24 & \underline{93.17} & \textbf{91.71} & \textbf{92.70} & \textbf{86.69} \\
			\hline
	\end{tabular}\label{tab:plist}
\end{table*}

\begin{figure*}[t]
\centering
{\includegraphics[width=0.7\linewidth]{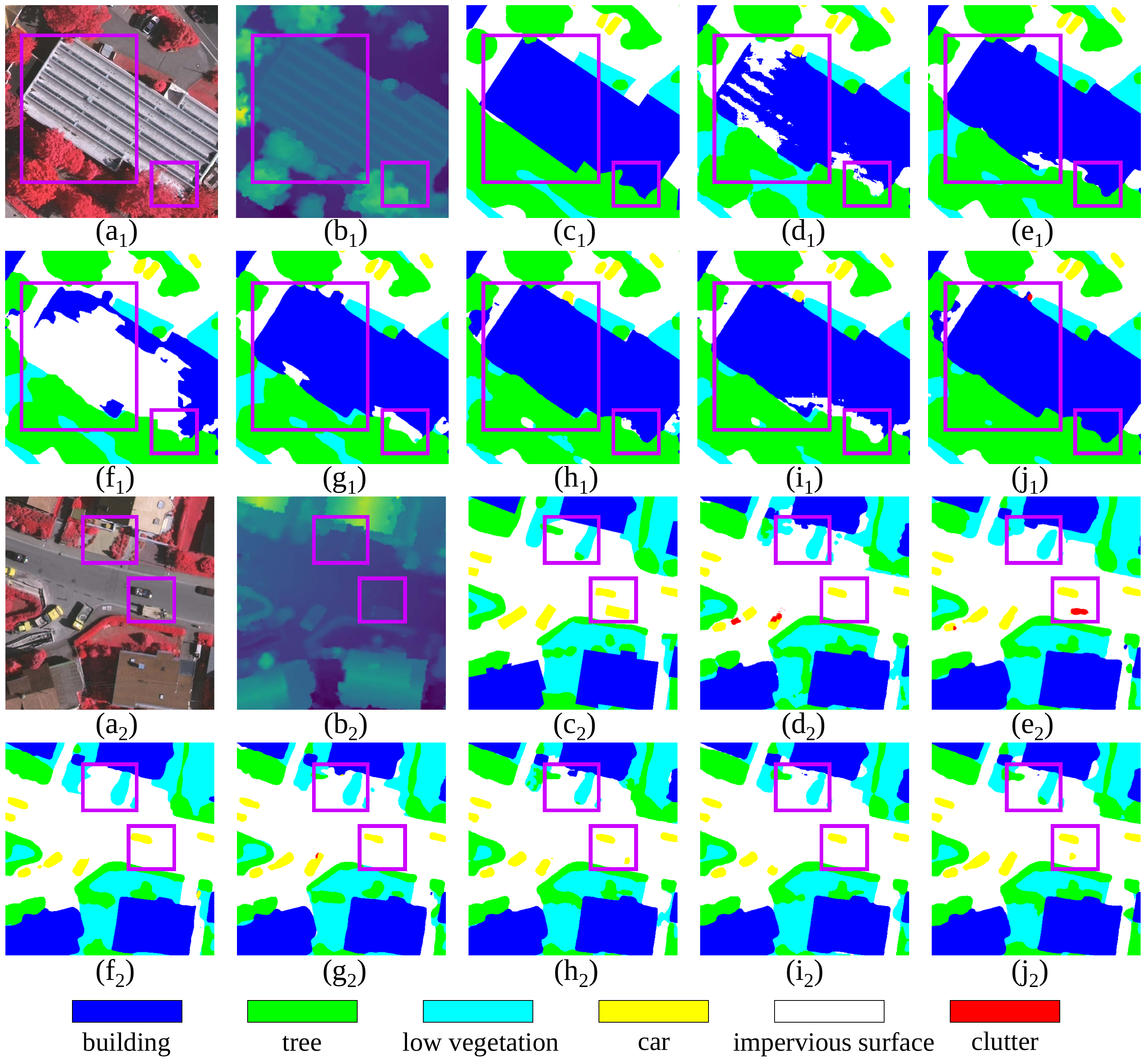}}
\caption{Visualized comparisons on the Vaihingen test set with the size of $512 \times 512$. (a) IRRG images, (b) DSM, (c) Ground Truth, (d) CMFNet, (e) FTransUNet, (f) MFTransNet, (g) CMGFNet, (h) FTransDeepLab, (i) MultiSenseSeg, (j) The proposed MFNet. To highlight the differences, some purple boxes are added to all subfigures.}
\label{fig_p_compare}
\label{fig_v_compare}
\end{figure*}

\begin{figure*}[t]
\centering
{\includegraphics[width=0.7\linewidth]{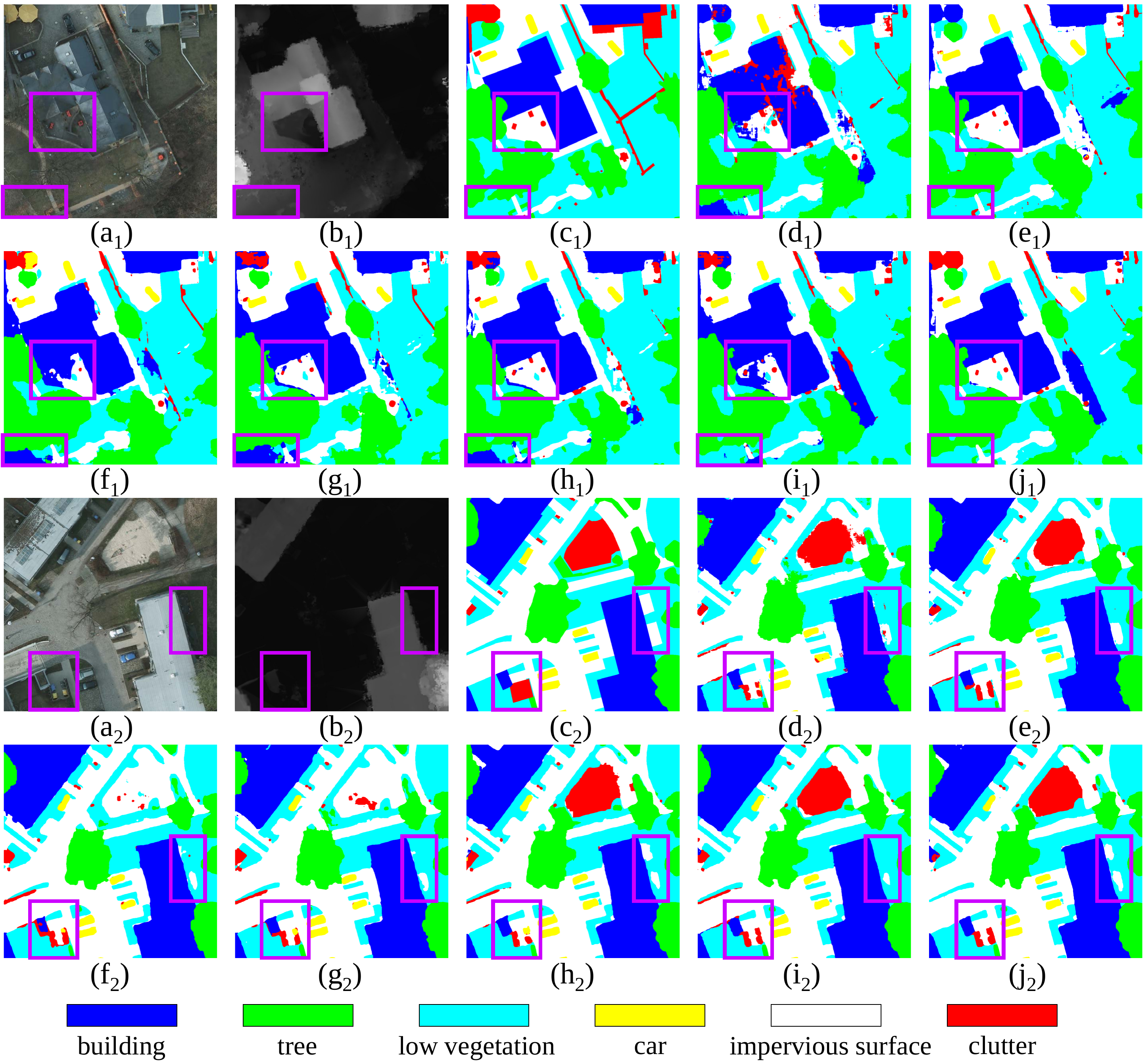}}
\caption{Visualized comparisons on the Potsdam test set with the size of $1024 \times 1024$. (a) IRRG images, (b) DSM, (c) Ground Truth, (d) CMFNet, (e) FTransUNet, (f) MFTransNet, (g) CMGFNet, (h) FTransDeepLab, (i) MultiSenseSeg, (j) The proposed MFNet. To highlight the differences, some purple boxes are added to all subfigures.}
\label{fig_p_compare}
\end{figure*}

\subsection{Implementation details}
All experiments were conducted using PyTorch on a single NVIDIA GeForce RTX 3090 GPU with 24GB of RAM. The stochastic gradient descent algorithm was used to train all the models under consideration with a learning rate of $0.01$, momentum of $0.9$, weight decay of $0.0005$, and a batch size of $10$. The batch size is reduced to $4$ when ViT-H is applied to meet the memory limit. All models were trained for a total of $50$ epochs, with each epoch comprising $1000$ batches. Basic data augmentation techniques, including random rotation and flipping, are applied after sample collection with the sliding window of size  $256 \times 256$. For MMAdapter, the down projection ratio is set to $0.25$. For MMLoRA, we follow the low-rank value in \citep{hu2021lora} as $4$. More details can be found in \href{https://github.com/sstary/SSRS}{https://github.com/sstary/SSRS}.

To assess the semantic segmentation performance on multimodal remote sensing data, we use overall accuracy (OA), mean F1 score (mF1), and mean intersection over union (mIoU). These standard metrics enable a fair comparison between the proposed MFNet and other state-of-the-art methods. Specifically, OA evaluates both foreground classes and the background class, while mF1 and mIoU are calculated for the five foreground classes.

\subsection{Performance Comparison}
We benchmarked the proposed MFNet against fifteen state-of-the-art methods, including PSPNet \citep{PSPNet}, MAResU-Net \citep{MAResUNet}, vFuseNet \citep{vfusenet}, FuseNet \citep{FuseNet}, ESANet \citep{ESANet}, SA-GATE \citep{SAGATE}, CMGFNet \citep{CMGFNet}, TransUNet \citep{Transunet}, CMFNet \citep{ma2022crossmodal}, UNetFormer \citep{wang2022unetformer}, MFTransNet \citep{he2023mftransnet}, FTransUNet \citep{ma2024ftransunet}, RS$^{3}$Mamba \citep{ma2024rs}, FTransDeepLab \citep{FTransDeepLab}, and MultiSenseSeg \citep{MultiSenseSeg}, most of which were specifically designed for remote sensing tasks. In our experiments, PSPNet, MAResU-Net, UNetFormer, and RS$^{3}$Mamba utilized only the optical images, to highlight the impact of DSM data and demonstrate the advantages of multimodal methods over single-modal ones. The other methods are state-of-the-art multimodal models based on different network architectures, including CNN, Transformer and Mamba. Taking into account the three different backbones provided by SAM, and the two multimodal fine-tuning architectures proposed in this work, we present six sets of experimental results for each dataset. The comparative quantitative results are presented in Table~\ref{tab:vlist} and Table~\ref{tab:plist}.

\subsubsection{Performance Comparison on the Vaihingen dataset}
As presented in Table~\ref{tab:vlist}, the proposed MFNet demonstrated substantial improvements in terms of OA, mF1 and mIoU metrics compared to existing segmentation methods. These results confirmed that our MFNet can effectively leverage the extensive general knowledge embedded in SAM. In particular, MFNet outperformed state-of-the-art models across four specific classes, namely {\em Building}, {\em Tree}, {\em Low Vegetation} and {\em Impervious surface}. In terms of the overall performance, the proposed MFNet (MMAdapter) with ViT-H achieved an OA of $92.97\%$, an mF1 score of $91.71\%$ and a mIoU of $85.03\%$, reflecting gains of $0.24\%$, $0.29\%$ and $0.50\%$ over the second best method MultiSenseSeg, respectively. Moreover, each of the three MFNet backbone variants offers unique advantages. Even the smallest variant, ViT-B, is comparable to most methods, further validating that our multimodal fine-tuning framework can efficiently utilize general knowledge from SAM to assist with the semantic segmentation of multimodal remote sensing data. This result demonstrated the practical value of the proposed MFNet and MMAdapter or MMLoRA in guiding the introduction of foundation models, like SAM, into multimodal remote sensing tasks. On the other hand, we observed that MFNet based on MMLoRA performed less effectively than the MMAdapter-based MFNet, which will be analyzed in Sec.~\ref{sec:complexity}.

Fig.~\ref{fig_v_compare} presents a visual comparison of the results produced by various methods and the best MFNet (MMAdapter) with ViT-H. MFNet demonstrates superior performance in generating sharper and more precise boundaries for ground objects, such as trees, cars, and buildings, resulting in clearer separations. This also helps preserve the integrity of the ground objects. Overall, the visualizations produced by MFNet exhibit a cleaner and more organized appearance. We attribute these improvements primarily to SAM's powerful feature extraction capabilities. By applying the multimodal fine-tuning mechanism, SAM's ability to segment every natural element is effectively extended to ground objects. 

\begin{table*}[t]\scriptsize
	\centering
	\caption{Quantitative results on the MMHunan dataset. The best results are in \textbf{Bold}. The second best results are \underline{underlined}. (\%).}
		\begin{tabular}{cccccccccccc}
			\hline
      \multirow{2}{*}{\textbf{Method}} & \multirow{2}{*}{\textbf{Backbone}} & \multicolumn{8}{c}{\textbf{OA}} & \multirow{2}{*}{\textbf{mF1}} & \multirow{2}{*}{\textbf{mIoU}} \\ \cline{3-10}
      & & \textbf{Cropland} & \textbf{Forest}  & \textbf{Grassland}  & \textbf{Wetland}   & \textbf{Water} & \textbf{Unused Land}  & \textbf{Built-up Area}  & \textbf{Total}  &                      &                       \\ \cline{1-1} \cline{11-12} 
			\hline
			FuseNet \citep{FuseNet} & VGG16 & 76.10 & 89.21 & 37.11 & 8.37 & 70.12 & 72.75 & 16.99 & 76.35 & 54.80 & 42.30 \\
			CMFNet \citep{ma2022crossmodal}  & VGG16 & 83.83 & 82.27 & \underline{43.88} & \underline{39.57} & 70.14 & \textbf{81.72} & 23.66 & 77.41 & 59.63 & 46.52 \\
			MFTransNet \citep{he2023mftransnet}  & ResNet34 & 78.99 & \underline{91.30} & 33.80 & 26.80 & 77.38 & 78.29 & 27.17 & 80.07 & 61.45 & 48.25 \\
			FTransUNet \citep{ma2024ftransunet} & R50-ViT-B & 78.75 & 90.54 & 32.13 & 27.51 & 76.64 & 75.59 & \textbf{48.63} & 79.47 & 61.95 & 48.78 \\
			CMGFNet \citep{CMGFNet} & ResNet34 & 82.08 & 87.90 & 37.50 & 26.48 & 79.82 & 74.64 & \underline{41.15} & 79.85 & 62.69 & 49.44 \\
			FTransDeepLab \citep{FTransDeepLab} & ResNet101 & \underline{79.39} & 88.89 & 35.71 & 30.88 & \underline{83.95} & 78.14 & 32.60 & 80.62 & 62.51 & 49.66 \\
			MultiSenseSeg \citep{MultiSenseSeg} & Segformer-B2 & 78.03 & 90.93 & 40.47 & 38.16 & 80.19 & \underline{81.03} & 38.14 & 80.51 & 63.74 & 50.76 \\
			\hline
			\multirow{3}{*}{MFNet (MMLoRA)}  & ViT-B & 76.83 & 90.69 & 24.16 & 22.03 & 80.17 & 78.12 & 29.66 & 79.35 & 58.79 & 46.54 \\
			  & ViT-L & \underline{79.39} & 88.89 & 35.71 & 30.88 & \underline{83.95} & 78.14 & 32.60 & 80.62 & 62.51 & 49.66 \\
				& ViT-H & 76.42 & 90.65 & 38.08 & 20.78 & 74.48 & 77.93 & 38.69 & 78.87 & 60.38 & 47.63 \\
			\hline
			\multirow{3}{*}{MFNet (MMAdapter)}  & ViT-B & 74.81 & \textbf{91.47} & 27.59 & 25.00 & \textbf{86.47} & 78.92 & 37.46 & 80.68 & 62.10 & 49.20 \\
			  & ViT-L & 79.19 & 89.52 & \textbf{46.07} & \textbf{42.23} & 81.15 & 77.58 & 39.65 & \underline{80.93} & \textbf{65.33} & \textbf{51.82} \\
				& ViT-H & \textbf{79.66} & 90.06 & 42.61 & 38.81 & 80.31 & 78.92 & 40.04 & \textbf{81.07} & \underline{64.13} & \underline{51.08} \\
			\hline
	\end{tabular}\label{tab:hlist}
\end{table*}

\begin{figure*}[t]
\centering
{\includegraphics[width=0.7\linewidth]{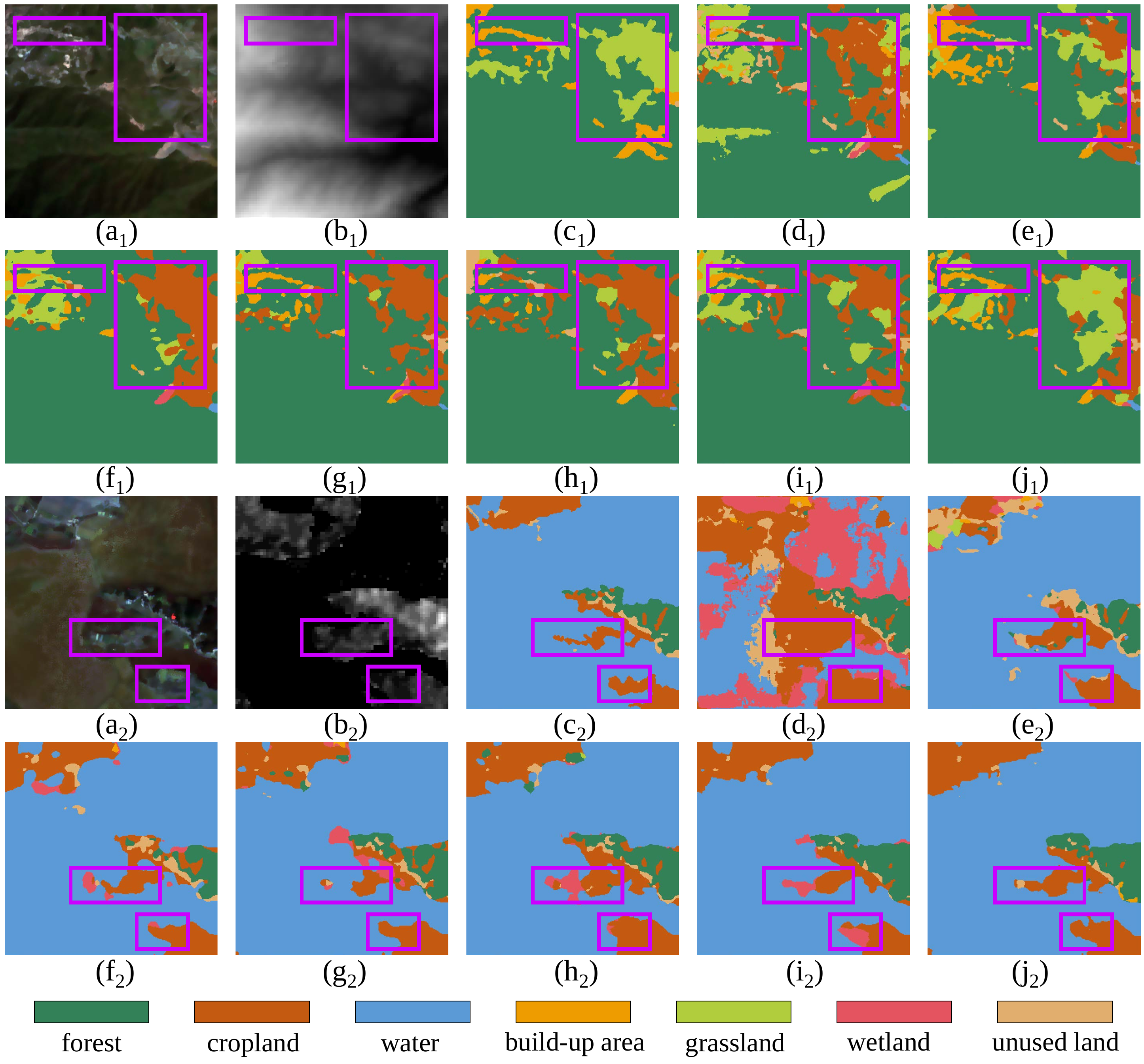}}
\caption{Visualized comparisons on the MMHunan test set with the size of $256 \times 256$. $1024 \times 1024$. (a) IRRG images, (b) DSM, (c) Ground Truth, (d) CMFNet, (e) FTransUNet, (f) MFTransNet, (g) CMGFNet, (h) FTransDeepLab, (i) MultiSenseSeg, (j) The proposed MFNet. To highlight the differences, some purple boxes are added to all subfigures.}
\label{fig_h_compare}
\end{figure*}

\subsubsection{Performance Comparison on the Potsdam dataset}
Experiments on the Potsdam dataset yielded results consistent with those from the Vaihingen dataset. As shown in Table~\ref{tab:plist}, the corresponding OA, mF1 and mIoU values of MFNet (MMAdapter) with ViT-H were $91.71\%$, $92.70\%$, $86.69\%$ respectively, which corresponds to increases of $0.37\%$, $0.29\%$ and $0.49\%$, respectively, over FTransUNet. Notably, significant gains were observed for {\em Building}, {\em Tree}, {\em Low Vegetation} and {\em Impervious Surface} compared to other state-of-the-art methods. The MFNet with smaller backbones also shows great performance. This flexibility allows MFNet to balance hardware requirements and performance needs across various application scenarios.

Fig.~\ref{fig_p_compare} shows a visualization example from Potsdam, where we observed more defined boundaries and intact object representations. These visual improvements are consistent with the mF1 and mIoU metrics shown in Table\ref{tab:plist}. Undoubtedly, it further validates the practical applicability of the proposed MFNet and MMAdapter or MMLoRA.

Additionally, it is observed that MFNet struggles to achieve optimal performance in identifying both {\em Tree} and {\em Low Vegetation} simultaneously. This challenge arises because both categories are characterized by irregular boundaries.Furthermore, their similarity, combined with their staggered or overlapping distribution, makes it challenging to distinguish between them. Combining SAM with more specialized designs to tackle the identification of these challenging categories presents an interesting future direction.

\subsubsection{Performance Comparison on the MMHunan dataset}
The experimental results on the MMHunan dataset are presented in Table~\ref{tab:hlist}. MFNet (MMAdapter) with a ViT-L backbone achieved OA, mF1, and mIoU scores of $80.93\%$, $65.33\%$, $51.82\%$, respectively, representing improvements of $0.42\%$, $1.59\%$ and $1.06\%$ over MultiSenseSeg. These results validate the consistent overall performance gains brought by our method. In addition, we observed an interesting phenomenon: across both fine-tuning strategies, ViT-H underperformed compared to ViT-L on this dataset. This suggests that in scenarios with small datasets, larger backbones may be more prone to overfitting, highlighting the importance of backbone selection for different remote sensing scenes. Fig.~\ref{fig_h_compare} shows a visualization example from MMHunan. In large-scale scenarios, the domain gap between remote sensing and natural images is more significant. However, our proposed approach successfully bridges this gap, allowing the vision foundation model's general understanding to be effectively transferred to remote sensing tasks, resulting in consistent performance improvements.

\begin{table*}[t]
	\centering
	\caption{Quantitative results on the Vaihingen dataset with different modalities and fine-tuning mechanisms. (\%).}
		\begin{tabular}{cccccccccc}
			\hline
      \multirow{2}{*}{\textbf{Modality}} & \multirow{2}{*}{\textbf{Fine-tuning}} & \multicolumn{6}{c}{\textbf{OA}} & \multirow{2}{*}{\textbf{mF1}} & \multirow{2}{*}{\textbf{mIoU}} \\ \cline{3-8}
      & & \textbf{Bui.}~ & \textbf{Tre.}  & \textbf{Low.}  & \textbf{Car}   & \textbf{Imp.}  & \textbf{Total}  &                      &                       \\ \cline{1-1} \cline{9-10} 
			\hline
			NIIRG & Without Adapter & 94.64 & 89.47 & 71.71 & 76.83 & 89.51 &  88.01 &  85.34   & 75.11  \\
			\hline
			NIIRG & Standard LoRA       & 96.50 & 93.62 & 80.35 & 86.32 & 92.78 & 92.00 & 90.35 & 82.74 \\
			NIIRG + DSM & Standard LoRA & 97.26 & 92.61 & 81.58 & 86.53 & 91.58 & 91.86 & 89.90 & 82.06 \\
			NIIRG + DSM & MFNet with MMLoRA & 96.85 & 92.89 & 81.09 & 89.95 & 93.28 & 92.22 & 91.09 & 83.96 \\
			\hline
			NIIRG & Standard Adapter       & 96.29 & 93.09 & 80.15 & 89.08 & 92.59 & 92.02 & 90.94 & 83.69 \\
			NIIRG + DSM & Standard Adapter & 99.02 & 91.68 & 83.04 & 89.71 & 92.90 & 92.80 & 91.30 & 84.35 \\
			NIIRG + DSM & MFNet with MMAdapter & 98.84 & 93.17 & 81.16 & 89.23 & 93.39 & 92.93 & 91.51 & 84.72 \\
			\hline	
	\end{tabular}\label{tab:modality}
\end{table*}

\subsection{Modality and Fine-tuning Analysis}
To illustrate the necessity of multimodal fine-tuning framework, we conducted the modality and fine-tuning analysis, with the results presented in Table~\ref{tab:modality}. In the first experiment, we only used single-modality data and did not apply any fine-tuning mechanism, while in the second and fifth experiments, we applied standard Adapter/LoRA mechanisms to fine-tune the SAM's image encoder. These experiments highlight the importance and need for multimodal information and fine-tuning mechanisms. In the third and sixth experiments, the SAM's image encoder retained standard Adapter/LoRA but excluded the proposed MMAdapter/MMLoR. Therefore, these experiments could still extract remote sensing multimodal features independently, but they lacked crucial information fusion during the encoding process. The fourth and seventh experiments contain multimodal information and the proposed MMAdapter/MMLoR.

Inspection of Table~\ref{tab:modality} first highlights the need for fine-tuning mechanisms. Without Adapter or LoRA, SAM struggles to effectively extract remote sensing features, leading to a significant decline in performance. Furthermore, the results reveal the great effectiveness of incorporating multimodal information. The improvement is particularly pronounced in the categories of {\em Building} and {\em Impervious surface}, which tend to have significant and stable surface elevation information. After that, this enhancement strengthens the model's ability to distinguish other categories. Additionally, we observe that the performance in the third experiment is lower than in the second. This is attributed to low-rank decomposition significantly reducing the dimensionality of task-specific information. Hereby, the heterogeneity between modalities complicates their fusion after the encoder. It highlights the challenge that improper fine-tuning poses in leveraging multimodal information. Our MMLoRA effectively addresses this challenge through a gradual feature fusion approach in the image encoder.

Overall, the integration of multimodal information provides comprehensive benefits across the board. The introduction of the MMAdapter and MMLoRA enables more effective utilization of DSM information, significantly enhancing the model's ability to extract and fuse multimodal information. As a result, the semantic segmentation performance is further improved.

\begin{figure*}[t]
\centering
{\includegraphics[width=0.9\linewidth]{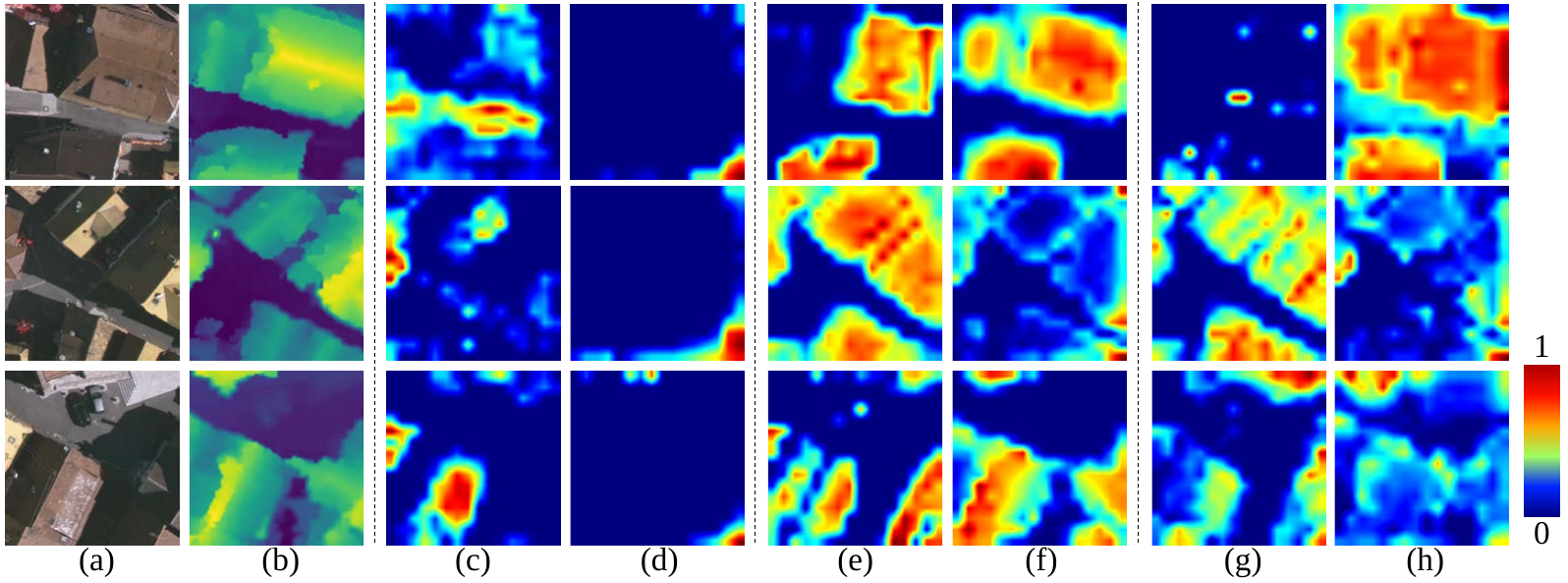}}
\caption{Four groups of heatmaps. (a) NIIRG images, (b) DSM, (c) heatmaps from NIIRG images and (d) heatmaps from DSM both generated by the original SAM, (e) heatmaps from NIIRG images and (f) heatmaps from DSM both generated by the proposed MMAdapter, (g) heatmaps from NIIRG images and (h) heatmaps from DSM both generated by the proposed MMLoRA. The high-value areas in the heatmaps indicate objects identified as buildings by the methods. The effectiveness of our MMAdapter and MMLoRA can be clearly observable.}
\label{fig_heatmaps}
\end{figure*}

\subsection{Ablation Study}
\subsubsection{Component Ablation}
The proposed MFNet consists of two core components: the SAM's image encoder with MMAdapter or MMLoRA and the DFM. To validate their effectiveness, ablation experiments were conducted by systematically removing specific components. As shown in Table~\ref{tab:ablalist}, two ablation experiments were designed. In the first experiment, the DFM was removed from MFNet, leading to a lack of deep analysis and integration of high-level abstract remote sensing semantic features. The second experiment follows the setup of the third and sixth experiments in Table~\ref{tab:modality}.

Before analyzing the ablation experiments, it is important to note that removing all Adapters or LoRAs from the SAM's image encoder severely degrades the model's performance, as confirmed in Fig.~\ref{fig_heatmaps}, which also illustrates the effectiveness of the multimodal fine-tuning mechanisms. Inspection of Fig.~\ref{fig_heatmaps}(c) and (d) reveals that SAM, without fine-tuning, cannot extract meaningful features from remote sensing data, rendering it unsuitable for semantic segmentation tasks. However, inspection of Fig.~\ref{fig_heatmaps}(d) and (f), or Fig.~\ref{fig_heatmaps}(g) and (h) shows that, after fine-tuning with the MMAdapter or MMLoRA, the heatmaps change dramatically. Furthermore, Fig.~\ref{fig_heatmaps}(f) and (h) clearly demonstrate that SAM, despite being trained on RGB optical images, is also effective when applied to non-optical DSM data. It is observed that DSM can effectively provide supplementary information. Therefore, the fine-tuned SAM's image encoder is capable of recognizing and segmenting remote sensing objects effectively in multimodal tasks.

\begin{table*}[h]
	\centering
	\caption{Ablation study of the proposed MFNet. The best results are in \textbf{Bold}.}
		\begin{tabular}{ccccc|ccccc}
			\hline
			\textbf{MMAdapter}   &  \textbf{DFM}    & \textbf{OA(\%)}   & \textbf{mF1(\%)}  & \textbf{mIoU(\%)}  & \textbf{MMLoRA}   &  \textbf{DFM}    & \textbf{OA(\%)}   & \textbf{mF1(\%)}  & \textbf{mIoU(\%)}   \\
			\hline
			\checkmark   &     &  92.73 &  91.23   & 84.25  & \checkmark  &  & 92.02 & 90.64 & 83.24 \\
		  &  \checkmark    &  92.80  &  91.30   & 84.35 & &  \checkmark    & 91.86 & 89.90 & 82.06  \\
			\checkmark   &  \checkmark    & \textbf{92.93}   & \textbf{91.51}   & \textbf{84.72} & \checkmark   &  \checkmark    & \textbf{92.22} & \textbf{91.09} & \textbf{83.96} \\
			\hline
	\end{tabular}\label{tab:ablalist}
\end{table*}

Inspection of Table~\ref{tab:ablalist} indicates that both the multimodal fine-tuning and DFM are essential for enhancing the performance of the proposed MFNet. Specifically, the MMAdapter and MMLoRA facilitate continuous information fusion, allowing for extraction and fusion of multimodal information as the encoding depth increases. The DFM verifies the importance of high-level features in the semantic segmentation of remote sensing data. In this work, we primarily introduce a new framework for leveraging SAM, rather than emphasize the high-level feature fusion techniques. Replacing DFM with a more advanced fusion model is expected to result in further performance improvement.

\begin{figure}[t]
\centering
{\includegraphics[width=0.9\linewidth]{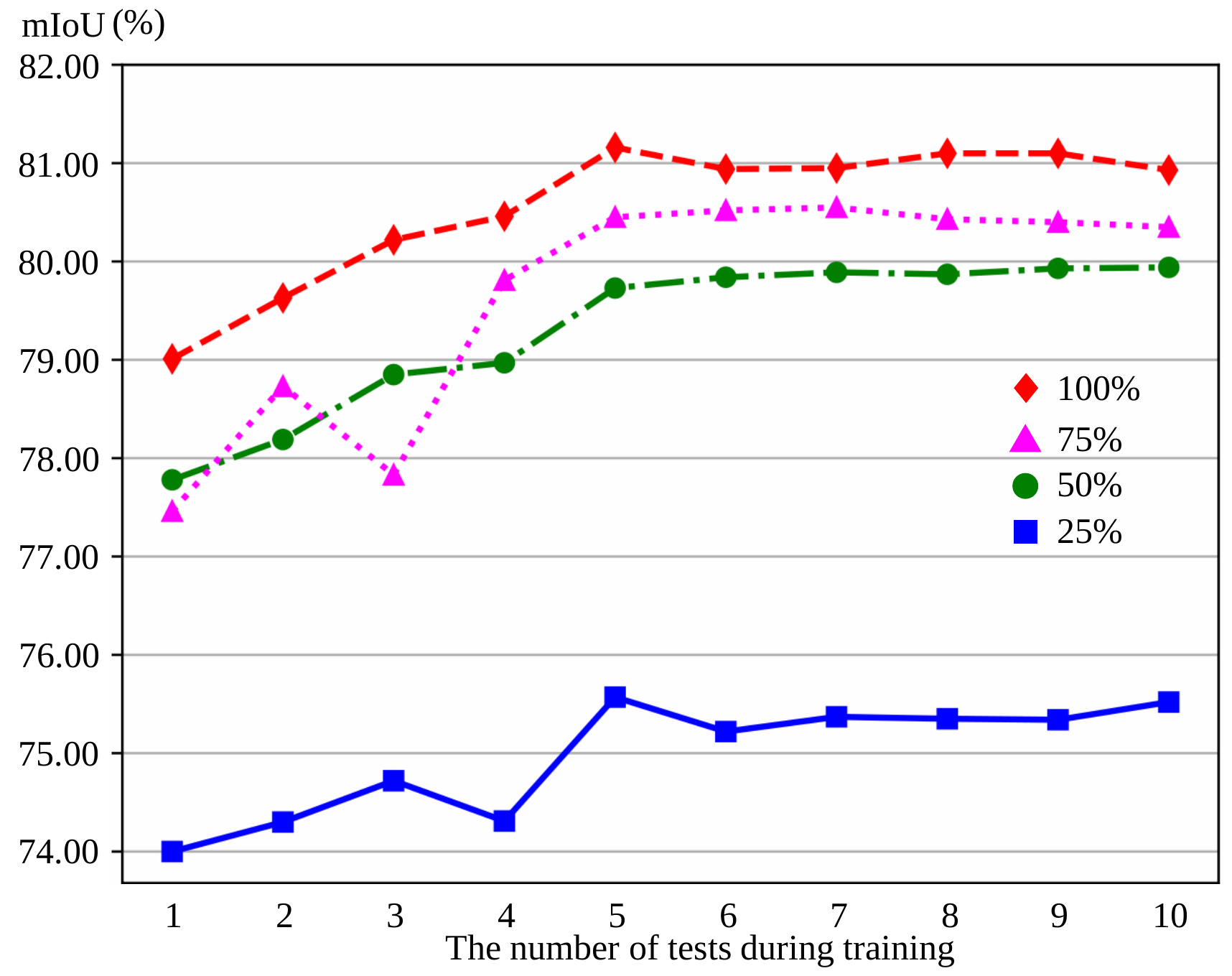}}
\caption{Relationship between training data volume and model performance during the training stage. As the model is unable to generate predictions in the absence of training data, we designed and carried out experiments with three levels of training data availability: 25\%, 50\%, and 75\%, in addition to using the complete training set (100\%).}
\label{dataamount}
\end{figure}

\subsubsection{Data Amount Ablation}
To investigate the fine-tuning efficiency of SAM on remote sensing tasks, we conducted experiments using varying proportions of the training data to explore the relationship between training data volume and model performance. Specifically, we fine-tuned the model using only 25\%, 50\%, and 75\% of the training set, while evaluating on the full test set. The results, shown in Fig.~\ref{dataamount}, reveal a phenomenon where data quantity plays a crucial role between 25\% and 50\%, yet the performance gains tend to saturate beyond the 50\% threshold. This suggests that SAM is capable of rapidly acquiring task-specific knowledge through fine-tuning, making further increases in training data yield diminishing returns in downstream performance. This finding offers valuable insights into the data requirements for related tasks, providing guidance on the efficient use of training data.

\begin{table}[h]
	\centering
	\caption{Model scale analysis measured by a $256 \times 256$ image on a single NVIDIA GeForce RTX 3090 GPU. For different MFNet configurations, the parameter statistics are: the fine-tuning parameters in SAM's image encoder + the parameters in DFM and the decoder. MIoU values are the results of the Vaihingen dataset. The best results are in \textbf{Bold}.}
		{\begin{tabular}{p{4.0cm}|p{1.2cm}p{1.0cm}p{1.0cm}}
			\hline
			\textbf{Method}   & \textbf{Parameter (M)} & \textbf{Memory (MB)}  &   \textbf{MIoU (\%)}  \\
			\hline
			PSPNet \citep{PSPNet}           & 46.72 & 3124 &  76.96\\
			MAResU-Net \citep{MAResUNet}    & 26.27 & 1908	&  79.89 \\
			UNetFormer \citep{wang2022unetformer} & 24.20	& 1980 &  81.97 \\
			RS$^{3}$Mamba \citep{ma2024rs}  & 43.32 & 1548 &  82.78 \\
			\hline
			TransUNet \citep{Transunet}     & 93.23 & 3028	&  78.26 \\
			FuseNet \citep{FuseNet}       & 42.08 & 2284 	&  78.71 \\
			vFuseNet \citep{vfusenet}      & 44.17 & 2618	&  78.92 \\
			ESANet \citep{ESANet}        & 34.03 & 1914	& 79.42 \\
			SA-GATE \citep{SAGATE}       & 110.85 & 3174	&  81.27 \\
			CMFNet \citep{ma2022crossmodal}       & 123.63 &  4058	&  81.44 \\
			MFTransUNet \citep{he2023mftransnet} & 43.77	& \textbf{1549} &  81.61\\
			CMGFNet  \citep{CMGFNet} &  64.20	& 2463 &  82.26 \\
			FTransUNet \citep{ma2024ftransunet}  &   160.88 & 3463 &  84.23 \\
			FTransDeepLab \citep{FTransDeepLab} & 69.86 & 1624 &  83.87 \\
			MultiSenseSeg \citep{MultiSenseSeg} & 60.46 & 2264 &  84.53 \\
			\hline
			MFNet (MMLoRA) (ViT-B)   &  	\textbf{1.03+6.22} &  1924 & 82.09 \\
			MFNet (MMLoRA) (ViT-L)   &  	2.75+6.22 &  4158 & 83.96 \\
			MFNet (MMLoRA) (ViT-H)   &  	4.59+6.22 &  6520 & 84.66 \\
			\hline
			MFNet (MMAdapter) (ViT-B)   &  	14.20+6.22 &  1872 & 83.24 \\
			MFNet (MMAdapter) (ViT-L)   &  	50.45+6.22 &  4242 & 84.72 \\
			MFNet (MMAdapter) (ViT-H)   &  	105.06+6.22 &  6854 & \textbf{85.03} \\
			\hline 
	\end{tabular}}\label{tab:scale}
\end{table}

\subsection{Model Scale Analysis}\label{sec:complexity}
The improved performance of MFNet is largely attributable to the general knowledge provided by the vision foundation model, SAM. However, SAM is also a large model, and the large model does not have advantages in terms of computational complexity or inference speed compared to existing general methods. Consequently, we focus on reporting the model's \emph{trainable} parameter number and memory footprint to measure its hardware requirements. 

Table~\ref{tab:scale} presents the results of the model scale for all methods compared in this work. As indicated in Table~\ref{tab:scale}, the proposed multimodal fine-tuning techniques allow the large foundation models to be used on a single GPU while maintaining a manageable number of trainable parameters and memory costs. The parameter statistics of MFNet are divided into two parts: the fine-tuning parameters in SAM's image encoder + the parameters in DFM and the decoder. The parameters in the latter remain consistent across different MFNet configurations. A comparison between MMLoRA and MMAdapter shows that MMLoRA significantly reduces the number of parameters by compressing thousands of dimensional spaces into a rank of $4$ through low-rank decomposition. While this approach is generally efficient, it may result in the loss of some essential information, especially when processing complex remote sensing data. Consequently, MMAdapter outperforms MMLoRA in terms of performance.

In our experiments, we successfully fine-tuned the ViT-L backbone on the same hardware with the same hyperparameters, achieving results that surpassed all existing methods. For the ViT-H backbone, we adjusted the batch size from $10$ to $4$ due to the GPU memory limitation. This reduction in batch size did not degrade performance, but further improved performance. These results prove the powerful feature extraction and fusion capabilities of the large vision foundation model. This work also offers valuable insights for exploring multimodal tasks with large models under constrained hardware conditions.

\subsection{Discussion}\label{sec:discussion}
This work introduces a unified multimodal fine-tuning framework with two SAM-based multimodal fine-tuning mechanisms. As an early exploration of this field, we thoroughly investigate the performance of the vision foundation model on remote sensing multimodal tasks by developing two classical fine-tuning approaches: Adapter and LoRA. Comprehensive analytical experiments are conducted to evaluate these methods. Additionally, MFNet offers a straightforward multimodal fusion network, paving the way for future research in the following directions:
\begin{itemize}[leftmargin=*]
\item{\bf Improving fine-tuning modules}: This work employs two representative fine-tuning techniques, namely Lora and Adapter, to demonstrate the framework’s effectiveness. Future research is encouraged to apply more advanced variants, such as \citep{lei2023conditional, chen2023hadamard, liu2024dora, hayou2024lora+} to diverse multimodal remote sensing tasks. In particular, more efficient fine-tuning strategies for large-scale models are worth exploring, as foundation models typically require substantial memory resources.

\item{\bf Improving fusion modules}: The work employs an adaptive weight-based method for feature fusion during the encoding stage. Future work could explore more advanced and effective fusion strategies tailored to MMAdapter and MMLoRA. Similarly, the fusion of deep high-level features can be enhanced with other fusion mechanisms, such as cross-attention, to improve performance.

\item{\bf Addressing challenging categories}: SAM's performance on challenging categories, such as distinguishing between very similar trees and low vegetation, or accurately detecting small objects such as cars, demands further investigation. To improve accuracy, it may be necessary to develop specialized object recognition modules that focus on these specific tasks. This could involve category-specific feature extraction techniques.

\item{\bf Exploration of other remote sensing modalities}: This study demonstrates the superiority of multimodal fine-tuning framework using optical images and DSM data as examples, and also provides valuable insights into the potential of combining these modalities. However, the performance of SAM on other remote sensing modalities, such as multispectral, LiDAR, and SAR, offers an exciting avenue for future investigation. Exploring these modalities could further enhance SAM’s capabilities and broaden its applicability in diverse remote sensing tasks.

\end{itemize}
Overall, this work serves as a foundational framework, with several aspects left open for further exploration. We hope it can be broadly extended to various types of multimodal remote sensing tasks.

\section{Conclusion}\label{sec:con}
In this study, we proposed a unified multimodal fusion framework with multimodal fine-tuning for remote sensing semantic segmentation, leveraging the general knowledge embedded in the vision foundation model, SAM. Using two representative single-modal fine-tuning mechanisms, namely Adapter and LoRA, we demonstrated the seamless integration of existing mechanisms into the proposed unified framework for extracting and fusing multimodal features from remote sensing data. The fused deep features were further refined using a pyramid-based DFM and reconstructed into segmentation maps. Comprehensive experiments on three benchmark multimodal datasets, ISPRS Vaihingen, ISPRS Potsdam and MMHunan, confirmed that MFNet achieves superior performance compared to current state-of-the-art segmentation methods. This research represents the first validation of SAM’s reliability on DSM data and offers a promising pathway for leveraging vision foundation models in multimodal remote sensing tasks. Moreover, the proposed framework has the potential to be extended to other remote sensing applications, including semi-supervised and unsupervised learning tasks.

\small
\bibliographystyle{IEEEtranN}
\bibliography{references}

@article{yao2023extended,
  title={Extended vision transformer ({ExViT}) for land use and land cover classification: A multimodal deep learning framework},
  author={Yao, Jing and Zhang, Bing and Li, Chenyu and Hong, Danfeng and Chanussot, Jocelyn},
  journal={IEEE Transactions on Geoscience and Remote Sensing},
  volume={61},
  pages={1--15},
  year={2023},
  publisher={IEEE}
}

@article{karmakar2023crop,
  title={Crop monitoring by multimodal remote sensing: A review},
  author={Karmakar, Priyabrata and Teng, Shyh Wei and Murshed, Manzur and Pang, Shaoning and Li, Yanyu and Lin, Hao},
  journal={Remote Sensing Applications: Society and Environment},
  pages={101093},
  year={2023},
  publisher={Elsevier}
}

@article{algiriyage2022multi,
  title={Multi-source multimodal data and deep learning for disaster response: a systematic review},
  author={Algiriyage, Nilani and Prasanna, Raj and Stock, Kristin and Doyle, Emma EH and Johnston, David},
  journal={SN Computer Science},
  volume={3},
  pages={1--29},
  year={2022},
  publisher={Springer}
}

@article{tf,
  title={Attention is all you need},
  author={Vaswani, Ashish and Shazeer, Noam and Parmar, Niki and Uszkoreit, Jakob and Jones, Llion and Gomez, Aidan N and Kaiser, {\L}ukasz and Polosukhin, Illia},
  journal={Advances in neural information processing systems},
  volume={30},
  pages = {1--11},
  year={2017}
}

@inproceedings{UNet,
 author = {Ronneberger, Olaf and Fischer, Philipp and Brox, Thomas},
 booktitle = {International Conference on Medical image computing and computer-assisted intervention},
 pages = {234--241},
 title = {{U-net}: Convolutional networks for biomedical image segmentation},
 year = {2015}
}

@article{li2022deep,
  title={Deep learning in multimodal remote sensing data fusion: A comprehensive review},
  author={Li, Jiaxin and Hong, Danfeng and Gao, Lianru and Yao, Jing and Zheng, Ke and Zhang, Bing and Chanussot, Jocelyn},
  journal={International Journal of Applied Earth Observation and Geoinformation},
  volume={112},
  pages={102926},
  year={2022},
  publisher={Elsevier}
}

@article{hong2021multimodal,
  title={Multimodal remote sensing benchmark datasets for land cover classification with a shared and specific feature learning model},
  author={Hong, Danfeng and Hu, Jingliang and Yao, Jing and Chanussot, Jocelyn and Zhu, Xiao Xiang},
  journal={ISPRS Journal of Photogrammetry and Remote Sensing},
  volume={178},
  pages={68--80},
  year={2021},
  publisher={Elsevier}
}

@article{wu2023medical,
  title={Medical {SAM} adapter: Adapting segment anything model for medical image segmentation},
  author={Wu, Junde and Ji, Wei and Liu, Yuanpei and Fu, Huazhu and Xu, Min and Xu, Yanwu and Jin, Yueming},
  journal={arXiv preprint arXiv:2304.12620},
  year={2023}
}

@ARTICLE{ma2024ftransunet,
  author={Ma, Xianping and Zhang, Xiaokang and Pun, Man-On and Liu, Ming},
  journal={IEEE Transactions on Geoscience and Remote Sensing}, 
  title={A Multilevel Multimodal Fusion Transformer for Remote Sensing Semantic Segmentation}, 
  year={2024},
  volume={62},
  number={},
  pages={1-15}
}

@article{wang2022unetformer,
  title={{UNetFormer}: A {UNet}-like transformer for efficient semantic segmentation of remote sensing urban scene imagery},
  author={Wang, Libo and Li, Rui and Zhang, Ce and Fang, Shenghui and Duan, Chenxi and Meng, Xiaoliang and Atkinson, Peter M},
  journal={ISPRS Journal of Photogrammetry and Remote Sensing},
  volume={190},
  pages={196--214},
  year={2022},
  publisher={Elsevier}
}

@article{li2021abcnet,
  title={{ABCNet}: Attentive bilateral contextual network for efficient semantic segmentation of Fine-Resolution remotely sensed imagery},
  author={Li, Rui and Zheng, Shunyi and Zhang, Ce and Duan, Chenxi and Wang, Libo and Atkinson, Peter M},
  journal={ISPRS journal of photogrammetry and remote sensing},
  volume={181},
  pages={84--98},
  year={2021},
  publisher={Elsevier}
}

@inproceedings{PSPNet,
  title={Pyramid scene parsing network},
  author={Zhao, Hengshuang and Shi, Jianping and Qi, Xiaojuan and Wang, Xiaogang and Jia, Jiaya},
  booktitle={Proceedings of the IEEE conference on computer vision and pattern recognition},
  pages={2881--2890},
  year={2017}
}

@article{MAResUNet,
  author={Li, Rui and Zheng, Shunyi and Duan, Chenxi and Su, Jianlin and Zhang, Ce},
  journal={IEEE Geoscience and Remote Sensing Letters}, 
  title={Multistage Attention ResU-Net for Semantic Segmentation of Fine-Resolution Remote Sensing Images}, 
  year={2022},
  volume={19},
  number={},
  pages={1-5},
}

@article{vfusenet,
 author = {Audebert, Nicolas and Le Saux, Bertrand and Lef{\`e}vre, S{\'e}bastien},
 journal = {ISPRS Journal of Photogrammetry and Remote Sensing},
 pages = {20--32},
 title = {Beyond {RGB}: Very high resolution urban remote sensing with multimodal deep networks},
 volume = {140},
 year = {2018}
}

@article{ResUNet-a,
 author = {Diakogiannis, Foivos I and Waldner, Fran{\c{c}}ois and Caccetta, Peter and Wu, Chen},
 journal = {ISPRS Journal of Photogrammetry and Remote Sensing},
 pages = {94--114},
 title = {{ResUNet-a}: A deep learning framework for semantic segmentation of remotely sensed data},
 volume = {162},
 year = {2020}
}

@inproceedings{FuseNet,
 author = {Hazirbas, Caner and Ma, Lingni and Domokos, Csaba and Cremers, Daniel},
 booktitle = {Asian conference on computer vision},
 pages = {213--228},
 title = {{FuseNet}: Incorporating depth into semantic segmentation via fusion-based cnn architecture},
 year = {2016}
}

@inproceedings{ESANet,
 author = {Seichter, Daniel and K{\"o}hler, Mona and Lewandowski, Benjamin and Wengefeld, Tim and Gross, Horst-Michael},
 booktitle = {2021 IEEE International Conference on Robotics and Automation (ICRA)},
 pages = {13525--13531},
 title = {Efficient rgb-d semantic segmentation for indoor scene analysis},
 year = {2021}
}

@inproceedings{SAGATE,
 author = {Chen, Xiaokang and Lin, Kwan-Yee and Wang, Jingbo and Wu, Wayne and Qian, Chen and Li, Hongsheng and Zeng, Gang},
 booktitle = {European Conference on Computer Vision},
 pages = {561--577},
 title = {Bi-directional cross-modality feature propagation with separation-and-aggregation gate for {RGB-D} semantic segmentation},
 year = {2020}
}

@article{CMGFNet,
  title={{CMGFNet}: A deep cross-modal gated fusion network for building extraction from very high-resolution remote sensing images},
  author={Hosseinpour, Hamidreza and Samadzadegan, Farhad and Javan, Farzaneh Dadrass},
  journal={ISPRS journal of photogrammetry and remote sensing},
  volume={184},
  pages={96--115},
  year={2022}
}

@article{Transunet,
  title={{TransUNet}: Transformers make strong encoders for medical image segmentation},
  author={Chen, Jieneng and Lu, Yongyi and Yu, Qihang and Luo, Xiangde and Adeli, Ehsan and Wang, Yan and Lu, Le and Yuille, Alan L and Zhou, Yuyin},
  journal={arXiv preprint arXiv:2102.04306},
  year={2021}
}

@ARTICLE{ma2022crossmodal,
  author={Ma, Xianping and Zhang, Xiaokang and Pun, Man-On},
  journal={IEEE Journal of Selected Topics in Applied Earth Observations and Remote Sensing}, 
  title={A Crossmodal Multiscale Fusion Network for Semantic Segmentation of Remote Sensing Data}, 
  year={2022},
  volume={15},
  number={},
  pages={3463-3474}
}

@article{he2023mftransnet,
  title={{MFTransNet}: A Multi-Modal Fusion with CNN-Transformer Network for Semantic Segmentation of {HSR} Remote Sensing Images},
  author={He, Shumeng and Yang, Houqun and Zhang, Xiaoying and Li, Xuanyu},
  journal={Mathematics},
  volume={11},
  number={3},
  pages={722},
  year={2023},
}

@ARTICLE{ma2024rs,
  author={Ma, Xianping and Zhang, Xiaokang and Pun, Man-On},
  journal={IEEE Geoscience and Remote Sensing Letters}, 
  title={{RS$^3$Mamba}: Visual State Space Model for Remote Sensing Image Semantic Segmentation}, 
  year={2024},
  volume={21},
  number={},
  pages={1-5}
}

@article{ma2024sam,
  author={Ma, Xianping and Wu, Qianqian and Zhao, Xingyu and Zhang, Xiaokang and Pun, Man-On and Huang, Bo},
  journal={IEEE Transactions on Geoscience and Remote Sensing}, 
  title={{SAM}-Assisted Remote Sensing Imagery Semantic Segmentation With Object and Boundary Constraints}, 
  year={2024},
  volume={62},
  pages={1-16},
}

@article{zhang2022multilevel,
  title={Multilevel deformable attention-aggregated networks for change detection in bitemporal remote sensing imagery},
  author={Zhang, Xiaokang and Yu, Weikang and Pun, Man-On},
  journal={IEEE Transactions on Geoscience and Remote Sensing},
  volume={60},
  pages={1--18},
  year={2022},
  publisher={IEEE}
}

@article{zhang2023cross,
  title={Cross-domain landslide mapping from large-scale remote sensing images using prototype-guided domain-aware progressive representation learning},
  author={Zhang, Xiaokang and Yu, Weikang and Pun, Man-On and Shi, Wenzhong},
  journal={ISPRS Journal of Photogrammetry and Remote Sensing},
  volume={197},
  pages={1--17},
  year={2023},
  publisher={Elsevier}
}

@article{vit,
 author = {Dosovitskiy, Alexey and Beyer, Lucas and Kolesnikov, Alexander and Weissenborn, Dirk and Zhai, Xiaohua and Unterthiner, Thomas and Dehghani, Mostafa and Minderer, Matthias and Heigold, Georg and Gelly, Sylvain and others},
 title = {An image is worth 16$\times$16 words: Transformers for image recognition at scale},
 journal={International Conference on Learning Representations},
 year={2021},
 pages = {1--22}
}

@inproceedings{kirillov2023segment,
  title={Segment anything},
  author={Kirillov, Alexander and Mintun, Eric and Ravi, Nikhila and Mao, Hanzi and Rolland, Chloe and Gustafson, Laura and Xiao, Tete and Whitehead, Spencer and Berg, Alexander C and Lo, Wan-Yen and others},
  booktitle={Proceedings of the IEEE/CVF International Conference on Computer Vision},
  pages={4015--4026},
  year={2023}
}

@article{yan2023ringmo,
  title={{RingMo-SAM}: A Foundation Model for Segment Anything in Multimodal Remote-Sensing Images},
  author={Yan, Zhiyuan and Li, Junxi and Li, Xuexue and Zhou, Ruixue and Zhang, Wenkai and Feng, Yingchao and Diao, Wenhui and Fu, Kun and Sun, Xian},
  journal={IEEE Transactions on Geoscience and Remote Sensing},
  volume={61},
  pages={1--16},
  year={2023},
  publisher={IEEE}
}

@article{pu2024classwise,
  title={Classwise-{SAM}-adapter: Parameter efficient fine-tuning adapts segment anything to sar domain for semantic segmentation},
  author={Pu, Xinyang and Jia, Hecheng and Zheng, Linghao and Wang, Feng and Xu, Feng},
  journal={arXiv preprint arXiv:2401.02326},
  year={2024}
}

@article{zhou2024mesam,
  title={{MeSAM}: Multiscale Enhanced Segment Anything Model for Optical Remote Sensing Images},
  author={Zhou, Xichuan and Liang, Fu and Chen, Lihui and Liu, Haijun and Song, Qianqian and Vivone, Gemine and Chanussot, Jocelyn},
  journal={IEEE Transactions on Geoscience and Remote Sensing},
  volume={62},
  pages={5623515},
  year={2024},
  publisher={IEEE}
}

@ARTICLE{ding2023adapting,
  author={Ding, Lei and Zhu, Kun and Peng, Daifeng and Tang, Hao and Yang, Kuiwu and Bruzzone, Lorenzo},
  journal={IEEE Transactions on Geoscience and Remote Sensing}, 
  title={Adapting Segment Anything Model for Change Detection in {VHR} Remote Sensing Images}, 
  year={2024},
  volume={62},
  number={},
  pages={1-11}
}

@article{mei2024scd,
  title={{SCD-SAM}: Adapting Segment Anything Model for Semantic Change Detection in Remote Sensing Imagery},
  author={Mei, Liye and Ye, Zhaoyi and Xu, Chuan and Wang, Hongzhu and Wang, Ying and Lei, Cheng and Yang, Wei and Li, Yansheng},
  journal={IEEE Transactions on Geoscience and Remote Sensing},
  year={2024},
  publisher={IEEE}
}

@article{mazurowski2023segment,
  title={Segment anything model for medical image analysis: an experimental study},
  author={Mazurowski, Maciej A and Dong, Haoyu and Gu, Hanxue and Yang, Jichen and Konz, Nicholas and Zhang, Yixin},
  journal={Medical Image Analysis},
  volume={89},
  pages={102918},
  year={2023},
  publisher={Elsevier}
}

@inproceedings{SAMRS,
  title={{SAMRS}: Scaling-up Remote Sensing Segmentation Dataset with Segment Anything Model},
  author={Di Wang and Jing Zhang and Bo Du and Minqiang Xu and Lin Liu and Dacheng Tao and Liangpei Zhang},
  booktitle={Thirty-seventh Conference on Neural Information Processing Systems Datasets and Benchmarks Track},
  year={2023},
}

@inproceedings{li2022exploring,
  title={Exploring plain vision transformer backbones for object detection},
  author={Li, Yanghao and Mao, Hanzi and Girshick, Ross and He, Kaiming},
  booktitle={European conference on computer vision},
  pages={280--296},
  year={2022},
  organization={Springer}
}

@INPROCEEDINGS{qi2024samp,
  author={Qi, Zipeng and Liu, Chenyang and Liu, Zili and Chen, Hao and Wu, Yongchang and Zou, Zhengxia and Shi, Zhenwei},
  booktitle={IGARSS 2024 - 2024 IEEE International Geoscience and Remote Sensing Symposium}, 
  title={{Multi-View} Remote Sensing Image Segmentation with {SAM} Priors}, 
  year={2024},
  volume={},
  number={},
  pages={8446-8449}
}

@article{chen2024change,
  title={Change Detection Between Optical Remote Sensing Imagery and Map Data via Segment Anything Model ({SAM})},
  author={Chen, Hongruixuan and Song, Jian and Yokoya, Naoto},
  journal={arXiv preprint arXiv:2401.09019},
  year={2024}
}

@inproceedings{houlsby2019parameter,
  title={Parameter-efficient transfer learning for {NLP}},
  author={Houlsby, Neil and Giurgiu, Andrei and Jastrzebski, Stanislaw and Morrone, Bruna and De Laroussilhe, Quentin and Gesmundo, Andrea and Attariyan, Mona and Gelly, Sylvain},
  booktitle={International conference on machine learning},
  pages={2790--2799},
  year={2019},
  organization={PMLR}
}

@inproceedings{
	hu2021lora,
	title={Lo{RA}: Low-Rank Adaptation of Large Language Models},
	author={Edward J Hu and yelong shen and Phillip Wallis and Zeyuan Allen-Zhu and Yuanzhi Li and Shean Wang and Lu Wang and Weizhu Chen},
	booktitle={International Conference on Learning Representations},
	pages={},
	year={2022}
}

@inproceedings{chen2022adaptformer,
  title={Adaptformer: Adapting vision transformers for scalable visual recognition},
  author={Chen, Shoufa and Ge, Chongjian and Tong, Zhan and Wang, Jiangliu and Song, Yibing and Wang, Jue and Luo, Ping},
  journal={Advances in Neural Information Processing Systems},
  volume={35},
  pages={16664--16678},
  year={2022}
}

@inproceedings{he2023parameter,
  title={Parameter-efficient model adaptation for vision transformers},
  author={He, Xuehai and Li, Chunyuan and Zhang, Pengchuan and Yang, Jianwei and Wang, Xin Eric},
  booktitle={Proceedings of the AAAI Conference on Artificial Intelligence},
  volume={37},
  number={1},
  pages={817--825},
  year={2023}
}

@inproceedings{liu2021swin,
 author = {Liu, Ze and Lin, Yutong and Cao, Yue and Hu, Han and Wei, Yixuan and Zhang, Zheng and Lin, Stephen and Guo, Baining},
 booktitle = {Proceedings of the IEEE/CVF International Conference on Computer Vision},
 pages = {10012--10022},
 title = {Swin transformer: Hierarchical vision transformer using shifted windows},
 year = {2021}
}

@article{ma2024segment,
  title={Segment anything in medical images},
  author={Ma, Jun and He, Yuting and Li, Feifei and Han, Lin and You, Chenyu and Wang, Bo},
  journal={Nature Communications},
  volume={15},
  number={1},
  pages={654},
  year={2024},
  publisher={Nature Publishing Group UK London}
}

@inproceedings{wang2024sam,
  title={{SAM}-clip: Merging vision foundation models towards semantic and spatial understanding},
  author={Wang, Haoxiang and Vasu, Pavan Kumar Anasosalu and Faghri, Fartash and Vemulapalli, Raviteja and Farajtabar, Mehrdad and Mehta, Sachin and Rastegari, Mohammad and Tuzel, Oncel and Pouransari, Hadi},
  booktitle={Proceedings of the IEEE/CVF Conference on Computer Vision and Pattern Recognition},
  pages={3635--3647},
  year={2024}
}

@inproceedings{zheng2020foreground,
  title={Foreground-aware relation network for geospatial object segmentation in high spatial resolution remote sensing imagery},
  author={Zheng, Zhuo and Zhong, Yanfei and Wang, Junjue and Ma, Ailong},
  booktitle={Proceedings of the IEEE/CVF conference on computer vision and pattern recognition},
  pages={4096--4105},
  year={2020}
}

@article{li2018deep,
  title={Deep learning for remote sensing image classification: A survey},
  author={Li, Ying and Zhang, Haokui and Xue, Xizhe and Jiang, Yenan and Shen, Qiang},
  journal={Wiley Interdisciplinary Reviews: Data Mining and Knowledge Discovery},
  volume={8},
  number={6},
  pages={e1264},
  year={2018},
  publisher={Wiley Online Library}
}

@article{ma2023unsupervised,
  title={Unsupervised domain adaptation augmented by mutually boosted attention for semantic segmentation of vhr remote sensing images},
  author={Ma, Xianping and Zhang, Xiaokang and Wang, Zhiguo and Pun, Man-On},
  journal={IEEE Transactions on Geoscience and Remote Sensing},
  volume={61},
  pages={1--15},
  year={2023},
  publisher={IEEE}
}

@article{gomez2015multimodal,
  title={Multimodal classification of remote sensing images: A review and future directions},
  author={G{\'o}mez-Chova, Luis and Tuia, Devis and Moser, Gabriele and Camps-Valls, Gustau},
  journal={Proceedings of the IEEE},
  volume={103},
  number={9},
  pages={1560--1584},
  year={2015},
  publisher={IEEE}
}

@article{li2024fusionsam,
  title={{FusionSAM}: Latent Space driven Segment Anything Model for Multimodal Fusion and Segmentation},
  author={Li, Daixun and Xie, Weiying and Cao, Mingxiang and Wang, Yunke and Zhang, Jiaqing and Li, Yunsong and Fang, Leyuan and Xu, Chang},
  journal={arXiv preprint arXiv:2408.13980},
  year={2024}
}

@inproceedings{
   zhang2023adaptive,
   title={Adaptive Budget Allocation for Parameter-Efficient {Fine-Tuning} },
   author={Qingru Zhang and Minshuo Chen and Alexander Bukharin and Pengcheng He and Yu Cheng and Weizhu Chen and Tuo Zhao},
   booktitle={The Eleventh International Conference on Learning Representations },
   year={2023},
	 pages={1--17}
}

@ARTICLE{lu2024multi,
  author={Lu, Xiaoyan and Weng, Qihao},
  journal={IEEE Transactions on Geoscience and Remote Sensing}, 
  title={{Multi-LoRA} Fine-Tuned Segment Anything Model for Urban Man-Made Object Extraction}, 
  year={2024},
  volume={62},
  number={},
  pages={1-19}
}

@article{ma2024frequency,
  author={Ma, Xianping and Zhang, Xiaokang and Ding, Xingchen and Pun, Man-On and Ma, Siwei},
  journal={IEEE Transactions on Geoscience and Remote Sensing}, 
  title={Decomposition-Based Unsupervised Domain Adaptation for Remote Sensing Image Semantic Segmentation}, 
  year={2024},
  volume={62},
  number={},
  pages={1-18}
}

@article{aghajanyan2020intrinsic,
  title={Intrinsic dimensionality explains the effectiveness of language model fine-tuning},
  author={Aghajanyan, Armen and Zettlemoyer, Luke and Gupta, Sonal},
  journal={arXiv preprint arXiv:2012.13255},
  year={2020}
}

@article{zhang2023customized,
  title={Customized segment anything model for medical image segmentation},
  author={Zhang, Kaidong and Liu, Dong},
  journal={arXiv preprint arXiv:2304.13785},
  year={2023}
}

@article{zhang2024asanet,
  title={{ASANet}: Asymmetric Semantic Aligning Network for RGB and SAR image land cover classification},
  author={Zhang, Pan and Peng, Baochai and Lu, Chaoran and Huang, Quanjin and Liu, Dongsheng},
  journal={ISPRS Journal of Photogrammetry and Remote Sensing},
  volume={218},
  pages={574--587},
  year={2024},
  publisher={Elsevier}
}

@article{lei2023conditional,
  title={{Conditional Adapters}: Parameter-efficient transfer learning with fast inference},
  author={Lei, Tao and Bai, Junwen and Brahma, Siddhartha and Ainslie, Joshua and Lee, Kenton and Zhou, Yanqi and Du, Nan and Zhao, Vincent and Wu, Yuexin and Li, Bo and others},
  journal={Advances in Neural Information Processing Systems},
  volume={36},
  pages={8152--8172},
  year={2023}
}

@inproceedings{chen2023hadamard,
  title={{Hadamard Adapter}: An extreme parameter-efficient adapter tuning method for pre-trained language models},
  author={Chen, Yuyan and Fu, Qiang and Fan, Ge and Du, Lun and Lou, Jian-Guang and Han, Shi and Zhang, Dongmei and Li, Zhixu and Xiao, Yanghua},
  booktitle={Proceedings of the 32nd ACM International Conference on Information and Knowledge Management},
  pages={276--285},
  year={2023}
}

@inproceedings{liu2024dora,
  title={{DoRA}: Weight-decomposed low-rank adaptation},
  author={Liu, Shih-Yang and Wang, Chien-Yi and Yin, Hongxu and Molchanov, Pavlo and Wang, Yu-Chiang Frank and Cheng, Kwang-Ting and Chen, Min-Hung},
  booktitle={Forty-first International Conference on Machine Learning},
  year={2024}
}

@article{hayou2024lora+,
  title={{LoRA+}: Efficient low rank adaptation of large models},
  author={Hayou, Soufiane and Ghosh, Nikhil and Yu, Bin},
  journal={arXiv preprint arXiv:2402.12354},
  year={2024}
}

@article{FTransDeepLab,
  title={{FTransDeepLab}: Multimodal Fusion Transformer-Based DeepLabv3+ for Remote Sensing Semantic Segmentation},
  author={Feng, Haixia and Hu, Qingwu and Zhao, Pengcheng and Wang, Shunli and Ai, Mingyao and Zheng, Daoyuan and Liu, Tiancheng},
  journal={IEEE Transactions on Geoscience and Remote Sensing},
  year={2025},
  publisher={IEEE}
}

@article{MultiSenseSeg,
  title={{MultiSenseSeg}: A cost-effective unified multimodal semantic segmentation model for remote sensing},
  author={Wang, Qingpeng and Chen, Wei and Huang, Zhou and Tang, Hongzhao and Yang, Lan},
  journal={IEEE Transactions on Geoscience and Remote Sensing},
  year={2024},
  publisher={IEEE}
}

@article{li2022dkdfn,
  title={{DKDFN}: Domain knowledge-guided deep collaborative fusion network for multimodal unitemporal remote sensing land cover classification},
  author={Li, Yansheng and Zhou, Yuhan and Zhang, Yongjun and Zhong, Liheng and Wang, Jian and Chen, Jingdong},
  journal={ISPRS Journal of Photogrammetry and Remote Sensing},
  volume={186},
  pages={170--189},
  year={2022},
  publisher={Elsevier}
}

\end{document}